\documentclass[letterpaper]{article} 
\usepackage{aaai24}  
\usepackage{times}  
\usepackage{helvet}  
\usepackage{courier}  
\usepackage[hyphens]{url}  
\usepackage{graphicx} 
\usepackage{natbib}  
\usepackage{caption} 
\usepackage{algorithm}
\usepackage{algorithmic}

\usepackage{newfloat}
\usepackage{listings}
\DeclareCaptionStyle{ruled}{labelfont=normalfont,labelsep=colon,strut=off} 
\floatstyle{ruled}
\newfloat{listing}{tb}{lst}{}
\floatname{listing}{Listing}
\usepackage{kotex}
\usepackage{cite}
\usepackage{adjustbox}
\usepackage{multirow}
\usepackage{amsmath}
\usepackage{amssymb}
\usepackage{booktabs}
\usepackage{mathtools}
\usepackage{amsthm}
\usepackage{algorithm, algorithmic}
\usepackage{enumitem}
\usepackage{multirow}
\usepackage{color}
\usepackage{comment}

\newcommand{\oursol}{\textnormal{SemTra}}
\newcommand{\TP}{\mathcal{P}}

\begin{document}

\author {
    Sangwoo Shin,
    Minjong Yoo,
    Jeongwoo Lee,
    Honguk Woo\thanks{Honguk Woo is the corresponding author.}
}
\affiliations {
    Department of Computer Science and Engineering,  Sungkyunkwan University\\
    jsw7460@skku.edu, mjyoo2@skku.edu, ljwoo98@skku.edu, hwoo@skku.edu
}

\title{SemTra: A Semantic Skill Translator for Cross-Domain Zero-Shot Policy Adaptation}

\maketitle

\begin{abstract}
This work explores the zero-shot adaptation capability of semantic skills, semantically interpretable experts' behavior patterns, in cross-domain settings, where a user input in interleaved multi-modal snippets can prompt a new long-horizon task for different domains.
In these cross-domain settings, we present a semantic skill translator framework $\oursol$  which utilizes a set of multi-modal models to extract skills from the snippets, and leverages the reasoning capabilities of a pretrained language model to adapt these extracted skills to the target domain. 
The framework employs a two-level hierarchy for adaptation: task adaptation and skill adaptation. 
During task adaptation, seq-to-seq translation by the language model transforms the extracted skills into a semantic skill sequence, which is tailored to fit the cross-domain contexts. Skill adaptation focuses on optimizing each semantic skill for the target domain context, through parametric instantiations that are facilitated by language prompting and contrastive learning-based context inferences. This hierarchical adaptation empowers the framework to not only infer a complex task specification in one-shot from the interleaved multi-modal snippets, but also adapt it to new domains with zero-shot learning abilities.
We evaluate our framework with Meta-World, Franka Kitchen, RLBench, and CARLA environments. The results clarify the framework's superiority in performing long-horizon tasks and adapting to different domains, showing its broad applicability in practical use cases, such as cognitive robots interpreting abstract instructions and autonomous vehicles operating under varied configurations.
\end{abstract}

\section{Introduction}
The promise of zero-shot policy deployment across different domains stems from the capability to immediately adapt its pre-trained knowledge to unfamiliar environments without the need for extensive data collection or fine-tuning. Such capability could revolutionize fields where safety is crucial, and where a single failure can lead to significant consequences and substantial costs, such as autonomous driving and robotics. 
However, achieving robust zero-shot adaptation in these fields remains challenging due to the intricacies of given tasks and the dynamic nature of their deployment environments~\cite{behdad_zeroshot_vehicle, garg_adapt_zeroshot}.

In the realm of reinforcement learning (RL) and imitation learning, policy adaptation has seen some exploration, with researchers leveraging various forms of task specification to deduce the given task domain. For instance, several studies utilized video demonstrations or expert trajectories from similar domains~\cite{mandi_mosaic, kuno_dail}, while others pivoted towards language instructions provided by users~\cite{goyal:retail, saycan}, or video-language interleaved demonstrations in multi-modal user interfaces~\cite{jiang2023vima}. Furthermore, in~\cite{promptdt}, expert trajectories from the target domain were directly employed for task inference.  

In this work, we investigate the problem of cross-domain zero-shot policy adaptation, with a focus on an inclusive and generalized approach to process user inputs relevant to long-horizon tasks across diverse domains. To be specific, we consider situations where the policy is prompted with a task-relevant input, presented as multi-modal interleaved snippets encompassing video, sensor data, and textual elements.  
In light of this consideration, we introduce a novel framework $\oursol$, grounded in the notion of semantic skill translation that spans multi-domains. The framework is designed with a two-tiered hierarchical adaptation process, first at the task level, followed by the skill level, to enhance the potential for zero-shot policy adaptation to target domains. 

During task adaptation, our framework harnesses the reasoning strengths of Pretrained Language Models (PLMs) to interpret the task specification~\cite{saycan, huang_inner_monologue}. This specification, encapsulated within a sequence of multi-modal interleaved snippets, is transformed into a sequence of semantically interpretable skill representations (semantic skills).
For skill adaptation, the framework employs a parametric approach to instantiate skills, integrating domain contexts into the semantic skills. These contexts may not be solely encapsulated in the task specification or input snippets, as they can also be dynamically captured during the evaluation time through interactions with the deployment environment. 
Subsequently, these domain contexts form domain-randomized meta-knowledge, enabling the framework to achieve context-parameterized skill instantiation. 

\color{black}

The contributions of our work are summarized as follows. 
    (i) We present the novel framework $\oursol$, addressing the challenging yet practical issues of cross-domain zero-shot policy adaptation for long-horizon tasks. 
    (ii) We devise a robust, hierarchical adaptation algorithm that leverages PLM prompting for the task-level adaptation and disentangles semantic skills and domain contexts to enable the skill-level optimization through parametric instantiations across different domains.
    (iii) We intensively evaluate our framework in over 200 cross-domain scenarios with Meta-World, Franka Kitchen, RLBench, and CARLA simulation environments, demonstrating its broad applicability in practical use cases, such as cognitive robots and autonomous driving.

\section{Problem Formulation}

We consider a multi-modal \textit{task prompt} $\TP := (x_1, \cdots, x_N)$ comprising a sequence of snippets, where each snippet $x_i$ is given in one of several modalities such as video, sensor data, and text. 
We presume that $\TP$ presents a source task specification, and some snippets may contain pertinent information or guidance beneficial for cross-domain adaptation. 
Given a single task prompt $\TP$, our work is to establish a framework to find the optimal model $\phi^*$. The model maps $\TP$ to a policy $\pi_{\TP}$, maximizing the expected evaluation performance:
\begin{equation}
    \label{eqn: semtra_objective} 
    \phi^* = \mathop{\textnormal{argmax}}_{\phi} \left[\underset{(D, \TP) \sim p_{\textnormal{tgt}}(D, \TP)}{\mathbb{E}} \left[\textnormal{E}_D(\phi(\TP))\right]\right].
\end{equation}
Here, $E_D(\phi(\mathcal{P}))$ is an evaluation function for the policy $\phi(\TP)$ in a target domain $D$.

\begin{figure}[t]
    \centering
    \includegraphics[width=0.40\textwidth]{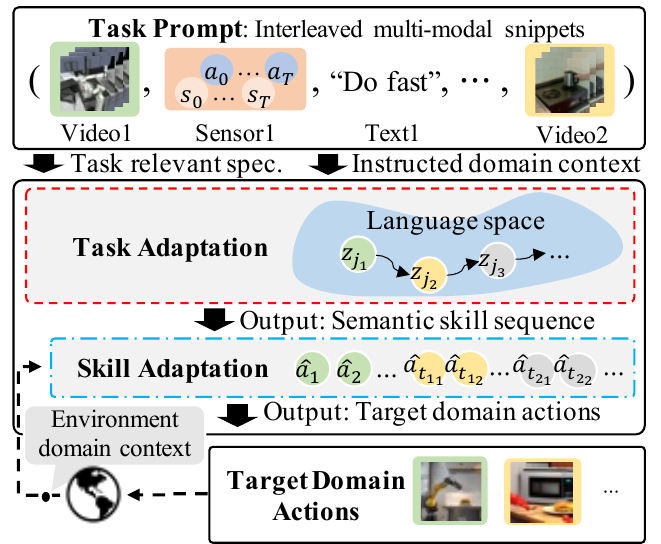}
    \caption{
    Cross-domain zero-shot adaptation for a multi-modal task prompt: our framework is given a task prompt filled with partial demonstrations and instructed contextual cues in multi-modal snippets. The framework conducts a two-phase adaptation, initially translating the snippets to semantic skills at the task level, and subsequently optimizing them into actions for the target domain at the skill level. 
    }
    \label{fig:overall}
\end{figure}

As illustrated in Figure 1, the task prompt is comprised of multiple modalities; e.g., the Video1 snippet offers a visual demonstration of an expert performing ``turning on the kitchen lights'', the Sensor1 snippet captures the trajectory of ``turning on the microwave.'', the Video2 snippet visually represents the process of ``boiling the water in the kettle'', and the Text1 snippet ``do fast'' contains contextual clues about the task execution in the environment, such as a fast-food restaurant. These snippets collectively formulate the source task specification and also serve as a guide for adaptation in the target domain. The evaluation function $E_D(\phi(\TP))$ in~\eqref{eqn: semtra_objective} assesses the task performance, particularly concerning cooking and serving maneuvers in the target domain $D$ with some stringent time constraints in this case. 

\begin{figure*}[t]
    \centering
        \includegraphics[width=1.0\textwidth]{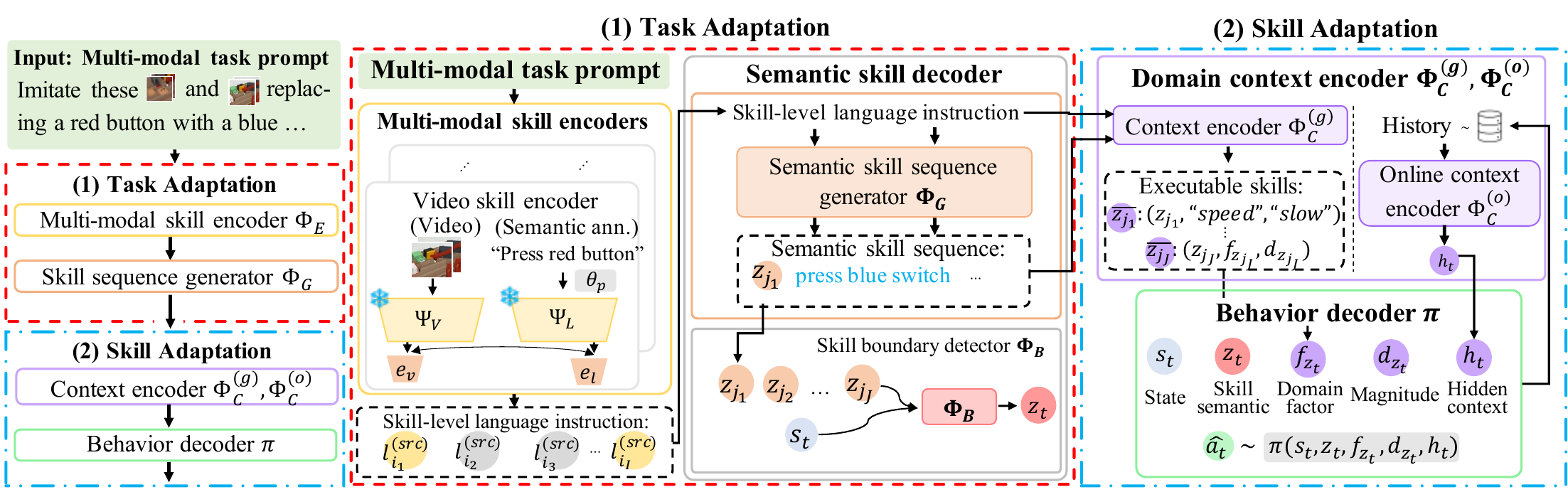}
    \caption{
    Two-phase policy adaptation in $\oursol$. (1) Task adaptation: the multi-modal skill encoders $\Phi_E$ produce a skill-level language instruction from the task prompt. In the figure, we specifically describe the training of a video skill encoder, contrastively learned through a pretrained VLM $(\Psi_V, \Psi_L)$. The skill-level instruction is then translated into a semantic skill sequence through the semantic skill sequence generator $\Phi_G$ based on a PLM. The skill boundary detector $\Phi_B$ infers the boundary of semantic skills upon a current state in the target domain. (2) Skill adaptation: the context encoder $\Phi^{(g)}_C$ identifies instructed domain contexts using both the skill-level instruction and the semantic skill sequence, generating an executable skill sequence. The online context encoder $\Phi^{(o)}_C$ captures environment hidden contexts at runtime. The behavior decoder $\pi$ generates actions optimized for the target domain, based on the executable skills and environment hidden contexts, along with the current state.
1    }
    \label{fig:methods}
\end{figure*}

\section{Approach}

\subsection{Overall Framework}
To tackle the challenge of cross-domain zero-shot adaptation with a multi-modal task prompt, we develop a semantic skill translator framework $\oursol$. 
As illustrated in Figure~\ref{fig:overall}, $\oursol$ decomposes the cross-domain adaptation hierarchically into two-phase linguistic adaptation procedures.
In the task adaptation phase, the framework captures the task discrepancy between the source and target domains by extracting a domain-invariant task solution using semantic skills. To do so, a pretrained vision-language model (VLM) is leveraged to establish the domain-agnostically learned semantic skills from the video snippets. The knowledge of PLM is also used to align the semantic skill sequence with the subtasks in the target domain. In the skill adaptation phase, the semantic skills are instantiated into an actual action sequence optimized for the target domain. 
\subsection{Task Adaptation}\label{sec: task interpretation}
For a task prompt $\TP = (x_1, \cdots x_N)$, our framework interprets it to produce a sequence of semantic skill embeddings $z_t$. These embeddings are represented in a language space 
and correspond to successive timesteps $t$ during task execution in the target environment.
\begin{equation}
    \label{eqn: task interpretation io}
    \Phi_D \circ \Phi_E: (s_t, \TP = (x_1, \cdots, x_N) ) \mapsto z_t
\end{equation}
This adaptation involves multi-modal skill encoders $\Phi_E$ and a semantic skill decoder $\Phi_D$, where each encoder takes the input specification as outlined by the task prompt, and the decoder maps the encoded specification into the semantic skills to be executed in the target environment. 
To train the encoder and decoder models, we use a dataset $\mathcal{D} = \{(\TP_i, \tau_i)\}_i$, where $\TP_i$ is a task prompt and $\tau$ is an expert trajectory in the target environment. Specifically, we use $\tau$ = $\{(s_t, a_t, v_t, l_t)\}_t$ with state $s_t$, action $a_t$, visual observation $v_t$, and semantic annotation $l_t$, where $l_t$ is a description of a skill or behavior at timestep $t$, e.g., ``push a button'', following~\cite{onis, saycan, pertsch:star, shridhar2022cliport}.

\label{sec: learning to semantic skills}
\subsubsection{Multi-modal skill encoders.}
The encoders $\Phi_E$ convert a task prompt into a skill-level language instruction $\eta$ as 
\begin{equation}
    \label{eqn: multi-modal skill encoder}
    \Phi_E: \TP = (x_1, \cdots, x_N) \mapsto \eta := (l^{(src)}_{i_1}, \cdots, l^{(src)}_{i_I})
\end{equation}
where each $l^{(src)}_{i_j}$ is a language token.
In this work, we consider three different modalities in snippets (i.e., video, sensor data, and text), although another modality can be incorporated through its respective encoder implementation.

To train the video skill encoder $\Psi_V$, which maps a video snippet $x = (v_{0:T})$ to skill-level language tokens, we leverage the video-to-text retrieval and text prompt learning with the pretrained VLM, V-CLIP~\cite{v-clip}, in a similar way of~\cite{onis}. Specifically, consider the V-CLIP $(\Psi_V, \Psi_L)$ model of a video encoder $\Psi_V$ and a text encoder $\Psi_L$. A text prompt $\theta_p$, a small set of learnable parameters in $\Psi_L$ is learned using a contrastive learning loss ~\cite{vinyals_infonce} on a batch $\{(v_{t-H:t}, l_t)\} \sim \mathcal{D}$. This optimization increases the similarity of video embedding $\Psi_V(v_{t-H:t})$ and its language embedding counterpart $\Psi_L(l_t; \theta_p)$. Once $\theta_p$ is trained, we use the retrieved semantic skill from the video snippet, i.e., $\Psi_V(v_{t-H:t}):= $
\begin{equation}
\label{eqn: video skill encoder}
    \textnormal{argmax}_{l \in \mathcal{L}_{vl}} \textnormal{sim}(\Psi_V(v_{t-H:t}), \Psi_L(l; \theta_p))
\end{equation}
\color{black}
where $\textnormal{sim} (\cdot, \cdot)$ is a similarity function and $\mathcal{L}_{vl} = \bigcup_{l \in \tau \in \mathcal{D}} \{l\}$ denotes a set of textually represented semantically interpretable skills. 
For sensor data snippets, we use a supervised classifier which predicts the semantic annotation $l_t$ from a state-action sub-trajectory $(s_{t-H': t}, a_{t-H': t})$. For text snippets, we use an identity function.

\subsubsection{Semantic skill decoder.}
Given a skill-level instruction $\eta$ in~\eqref{eqn: multi-modal skill encoder} and state $s_t$, the semantic skill decoder $\Phi_D$ yields a semantic skill $z_t$  for each timestep $t$, i.e., 
\begin{equation}
    \Phi_D: (s_t, \eta) \mapsto z_t
    \label{eqn: semantic skill decoder}
\end{equation}
by using two models, the semantic skill sequence generator $\Phi_G$ and the skill boundary detector $\Phi_B$.
The former $\Phi_G$ generates a sequence of target semantic skill embeddings through seq-to-seq translation, leveraging the knowledge of PLMs.
\begin{equation}
    \label{sequence generator} 
    \Phi_G: \eta \mapsto (z_{j_1}, \cdots, z_{j_J})
\end{equation}
At each timestep $t$, the latter $\Phi_B$ determines whether to continue executing the current skill $z_t$ or to transition to the next skill in the sequence until the task is completed. It evaluates the probability that ongoing semantic skill is terminated:
\begin{equation}
    \label{eqn: boundary detector}
    \Phi_B: (s_t, z_t, s_{t_0}) \mapsto p \in [0, 1]
\end{equation}
with state $s_t$, semantic skill $z_t$, and its initial state $s_{t_0}$.

For training $\Phi_G$, we exploit the skill encoders $\Phi_E$ that can extract a target semantic sequence $(z_{j_1}, \cdots, z_{j_J})$ from the video trajectory of $\tau^{(tgt)}$ through~\eqref{eqn: video skill encoder}. Subsequently, $\Phi_G$ is supervised using a cross-entropy-based loss on these target semantic sequences. Regarding $\Phi_B$, it is trained using a binary cross-entropy loss, i.e., $\textnormal{BCE} (\Phi_B(s_t, z_t, s_{t_0}), 1_{z_{t} \neq z_{t+1}})$. 
This task adaptation procedure is presented in the middle of Figure~\ref{fig:methods}.

\begin{algorithm}[t]
\fontsize{9.5pt}{11pt}\selectfont
\caption{Two-phase adaptation of $\oursol$}
\begin{algorithmic}[1]
    \STATE Task prompt $\mathcal{P}$, target trajectory $\tau^{(tgt)} = \{(s_t, v_t, a_t, l_t)\}$
    \STATE Domain contexts $\{\bar{z}'_{j_1}, \cdots, \bar{z}'_{j_J}\}$ of $\tau^{(tgt)}$
    \STATE Multi-modal skill enc. $\Phi_E$, skill sequence generator $\Phi_G$
    \STATE Skill boundary detector $\Phi_B$, context encoder $\Phi^{(g)}_C$
    \STATE Online context encoder $\Phi^{(o)}_C$, behavior decoder $\pi$
    
    \nonumber \textit{\textbf{/* Task adaptation in Section~\ref{sec: task interpretation} */}}
    \STATE $\eta \leftarrow \Phi_E(\mathcal{P})$ using~\eqref{eqn: multi-modal skill encoder}, $\xi \leftarrow \Phi_G(\eta)$ using~\eqref{sequence generator}
    \STATE $\xi' \leftarrow \{ \Psi_V(v_{t-H: t}): t \}$ with deduplication using~\eqref{eqn: video skill encoder}
    \STATE $\Phi_G \leftarrow \Phi_G - \nabla_{\Phi_G} (\textnormal{CrossEntropy}(\xi, \xi'))$
    \STATE $(z_0, \cdots, z_T) \leftarrow \{ \Psi_V(v_{t-H: t}): t \}$ using~\eqref{eqn: video skill encoder}
    \STATE $\Phi_B \leftarrow \Phi_B - \nabla_{\Phi_B}(\textnormal{BCE}(\Phi_B(s_t, z_t, s_{t_0}), 1_{z_t \neq z_{t+1}}))$
    
    \nonumber \textit{\textbf{/* Skill adaptation in Section~\ref{sec: skill interpretation} */}}
    \STATE $\{ \bar{z}_{j_1}, \cdots, \bar{z}_{j_J} \} \leftarrow \Phi^{(g)}_C(\eta, \xi)$ using~\eqref{eqn: context encoder}
    \STATE $\Phi^{(g)}_C \leftarrow \Phi^{(g)}_C - \nabla_{\Phi^{(g)}_C}(\textnormal{CrossEntropy}(\bar{z}_{j_i}, \bar{z}'_{j_i}))$
    \STATE $h_t \leftarrow \Phi^{(o)}_C(s_{t-H': t}, a_{t-H': t-1})$ using~\eqref{eqn: online context encoder}
    \STATE $\pi \leftarrow \pi - \nabla_{\pi}(\textnormal{MSE}(\pi(s_t, \bar{z}_t, h_t), a_t)$
\end{algorithmic}
\label{alg: train}
\end{algorithm}

\subsection{Skill Adaptation}\label{sec: skill interpretation}
In conjunction with the task adaptation, which generates a semantic skill sequence, the framework also encompasses a skill adaptation phase, where each skill is adapted individually to suit the target domain. It takes as input the semantic skill sequence $\xi := (z_{j_1}, \cdots, z_{j_J})$ in~\eqref{sequence generator}, paired with a skill-level language instruction $\eta$ in~\eqref{eqn: multi-modal skill encoder}, and the current state $s_t$. It continuously generates an action $a_t$ for each timestep during the execution of $\xi$ in the target domain, employing a context encoder $\Phi^{(g)}_C$ and a behavior decoder $\pi$.
\begin{equation}
    \pi \circ \Phi^{(g)}_C : (s_t, \eta, \xi) \mapsto a_t
\end{equation}

\subsubsection{Domain context encoder.}
The context encoder $\Phi^{(g)}_C$ transforms a semantic skill sequence into a series of executable skills appropriate for the target domain, by taking into account the cross-domain contexts encapsulated in a given task prompt. This involves utilizing the knowledge of PLMs to capture domain-specific features from the coupled skill-level instruction $\eta$ and its associated semantic skill sequence $\xi$.
\begin{equation}
    \Phi^{(g)}_C: (\eta, \xi) \mapsto (\bar{z}_{j_1}, \cdots, \bar{z}_{j_J}), \ \text{where}  \ \bar{z} := (z, f_z, d_z)
    \label{eqn: context encoder}
\end{equation}
Here, each executable skill $\bar{z}$ is parameterized by a semantic skill $z \in \xi$, a domain factor $f_z$, and a magnitude $d_z$ associated with $f_z$. The cross-domain context information in $\eta$ is used to infer the domain factor, which modulates the actual action execution of each semantic skill according to the specific requirements of the target domain; e.g., the domain factor for the semantic skill ``move to a destination''  could be the surface type or the required speed in the target domain.

\subsubsection{Behavior decoder.}
Given an executable skill sequence $(\bar{z}_{j_1}, \cdots, \bar{z}_{j_J})$ in~\eqref{eqn: context encoder}, the behavior decoder $\pi$ is learned to infer an action sequence optimized for the target domain.
\begin{equation}
    \pi: (s_t, \bar{z_t}) \mapsto a_t
    \label{eqn: behavior decoder}
\end{equation}
Furthermore, to adapt to temporal domain shifts that are not fully covered by the task prompt, an online context encoder $\Phi^{(o)}_C$ is utilized to infer a hidden context $h_t$ over timesteps $t$:
\begin{equation}
    \label{eqn: online context encoder}
    \Phi^{(o)}_C: (s_{t-H': t}, a_{t-H': t-1}) \mapsto h_t.
\end{equation}
Then, this hidden context is concatenated with $\bar{z}_t$ in~\eqref{eqn: behavior decoder}. This $\Phi^{(o)}_C$ can be implemented by using reconstruction~\cite{varibad, finn_pearl} or contrastive learning~\cite{onis, ood_imitation}. This skill adaptation procedure is on the right side of Figure~\ref{fig:methods}.
Algorithm~\ref{alg: train} outlines the two-phase adaptation in $\oursol$.

\section{Evaluation}
\subsection{Experiment Setting}
\subsubsection{Environments.}
For cross-domain adaptation evaluations, we use the Franka Kitchen~\cite{d4rl} (FK) and Meta-World~\cite{metaworld} (MW) environments. 
In these environments, each task $\mathcal{T}$ is composed of $N$ subtasks that must be completed in the correct order. This allows us to explore the compositionality of complex tasks and investigate the long-horizon imitation performance~\cite{onis, simpl, spirl, gupta_rpl}.

\subsubsection{Baselines.}
For comparison, we test \textbf{VIMA}~\cite{jiang2023vima} that incorporates multi-modal task prompts to a cross-attention-based Transformer for one-shot imitation learning, and we implement \textbf{TP-GPT}, a multi-modal task prompting approach, that specifically uses the decoder-only Transformer with self-attention~\cite{behavior_dt, promptdt}. Furthermore, we implement a video generation-based task prompting method \textbf{VCP}, similar to~\cite{Yulin_updp, elliot_vip} that incorporates task prompts to video embeddings-conditioned policies. 
We also implement a language generation-based task prompting method \textbf{TP-BCZ} that employs the state-of-the-art multi-modal zero-shot imitation framework~\cite{bc-z}. 

\subsubsection{Evaluation metrics.}
We employ two performance evaluation metrics in cross-domain zero-shot policy adaptation: 
\textbf{$K$-rate} (KR) that evaluates the rate of unseen task domains in which at least $K$ out of $N$ subtasks are completed, similar to the evaluation metric used in long-horizon tasks by~\cite{calvin};
\textbf{$N$-rate} (NR) that evaluates the average rate of successfully completed subtasks out of $N$ subtasks. 
In our evaluation, we set $N=4$ for Franka Kitchen and $N=3$ for Meta-World.

\subsubsection{Evaluation datasets.}
We use 9 unseen long-horizon evaluation tasks (e.g., each being defined by an unseen order of subtasks) in Franka Kitchen and 8 unseen long-horizon tasks in Meta-World. 
We also use 12 skill-level domain contexts (e.g., instructed domain contexts for time constraints and time-varying domain contexts for the non-stationarity of environments). 
Based on the combination of these unseen tasks and contexts, we establish a total of 108 and 96 distinct unseen target domains for Franka Kitchen and Meta-World, respectively. 
Detailed explanations for training datasets and examples of task prompts can be found in Appendix B.

\subsection{Results}

\subsubsection{Cross-domain performance.}
We evaluate the cross-domain performance in both $K$-rate and $N$-rate, wherein a policy is evaluated in a zero-shot manner for different target domains. In Table~\ref{exp:cross-domain}, $\oursol$ consistently outperforms the baselines, specifically $36.34\%$ and $66.24\%$ $N$-rate higher than the most competitive baselines VCP in Franka Kitchen and TP-GPT in Meta-World, respectively.
With respect to the $K$-rate metric, the margin of superiority becomes more significant for larger $K$, underlining the robust adaptation capabilities of $\oursol$, especially for long-horizon tasks.

\color{black}
\begin{table}[h]
    \small
    \centering
    {
    \begin{tabular}{c|c||c|c|c|c|c}
    \specialrule{.1em}{.05em}{.05em}
    \textbf{Env} & \textbf{KR} & \textbf{VIMA} & \textbf{BCZ} & \textbf{GPT} & \textbf{VCP} & \textbf{SemTra} \\
    \specialrule{.1em}{.05em}{.05em}

    \multirow{5}{*}{FK} 
    & $K$=1 & 63.6 & 55.1 & 51.1 & 62.2 & \textbf{100.0} \\ 
    & $K$=2 & 31.3 & 19.5 & 32.0 & 32.8 & \textbf{80.0} \\ 
    & $K$=3 & 13.9 & 3.1 & 14.2 & 16.8 & \textbf{61.7}  \\ 
    & $K$=4 & 5.3 & 0.0 & 7.5 & 7.5 & \textbf{23.1} \\ 
    \cline{2-7}
    & \textbf{NR} & 28.3 & 19.4 & 26.2 & 29.8 & \textbf{66.2} \\
    \specialrule{.1em}{.05em}{.05em} 

    \multirow{4}{*}{MW} 
    & $K$=1 & 18.5 & 17.4 & 35.4 & 22.7 & \textbf{86.2}  \\ 
    & $K$=2 & 4.8 & 6.4 & 3.2 & 4.8 & \textbf{82.8} \\ 
    & $K$=3 & 0.0 & 0.0 & 0.0 & 0.5 & \textbf{72.6} \\ 
    \cline{2-7}
    & \textbf{NR} & 7.6 & 7.8 & 13.1 & 8.4 & \textbf{79.3} \\
    \specialrule{.1em}{.05em}{.05em} 
    
    \end{tabular}
    }
    \caption{
    Cross-domain performance in $K$-rate and $N$-rate.
    }
    \label{exp:cross-domain}
\end{table}

\subsubsection{Single-domain (reference) performance.}
To analyze the cross-domain performance, we also use the single-domain scenarios (i.e., one-shot imitation within a single-domain) as a reference, where we treat the task prompt as a demonstration specific to the target domain. As shown in Table~\ref{exp:single-domain}, $\oursol$ shows $13.89\%$ and $58.33\%$ higher $N$-rate than the most competitive baseline TP-GPT in Franka Kitchen and Meta-World, respectively. More importantly, we confirm the robust cross-domain performance of $\oursol$ (in Table~\ref{exp:cross-domain}), which is comparable to this single-domain performance; specifically, the degradation is minimal, when transitioning from single-domain to cross-domain experiments. In contrast, the baselines experience a relatively large degradation.

\begin{table}[h]
    \small
    \centering
    {
    \begin{tabular}{c|c||c|c|c|c|c}
    \specialrule{.1em}{.05em}{.05em}
    \textbf{Env} & \textbf{KR} & \textbf{VIMA} & \textbf{BCZ} & \textbf{GPT} & \textbf{VCP} & \textbf{SemTra} \\
    \specialrule{.1em}{.05em}{.05em}

    \multirow{5}{*}{FK} 
    & $K$=1 & 96.2 & 62.9 & 92.5 & 77.7 & \textbf{100.0} \\ 
    & $K$=2 & 62.9 & 22.2 & 74.0 & 55.5 & \textbf{85.1} \\ 
    & $K$=3 & 22.2 & 3.7 & 40.7 & 29.6 & \textbf{66.6} \\ 
    & $K$=4 & 11.1 & 0.0 & 19.6 & 11.1 & \textbf{40.7} \\ 
    \cline{2-7}
    & \textbf{NR} & 48.1 & 22.2 & 59.2 & 43.5 & \textbf{73.1} \\
    \specialrule{.1em}{.05em}{.05em} 

    \multirow{4}{*}{MW} 
    & $K$=1 & 37.5 & 29.1 & 33.3 & 50.0 & \textbf{100.0} \\ 
    & $K$=2 & 12.5 & 16.6 & 8.3 & 12.5 & \textbf{79.1} \\ 
    & $K$=3 & 3.7 & 1.5 & 0.0 & 0.0 & \textbf{62.5} \\ 
    \cline{2-7}
    & \textbf{NR} & 16.6 & 15.2 & 22.2 & 20.8 & \textbf{80.5}  \\
    \specialrule{.1em}{.05em}{.05em} 
    
    \end{tabular}
    }
    \caption{
    Single-domain performance in $K$-rate and $N$-rate.
    }
    \label{exp:single-domain}
\end{table}
In the following, we use this single-domain performance in $N$-rate as a reference in comparison. 
\subsubsection{Task-level cross-domain performance.}
We specifically investigate the cross-domain adaptation performance under task-level domain changes, the scenarios where the order of subtasks in the target domain is not the same as that of a given task prompt. For the cross-domain setting, we consider task prompts involving snippets beneficial for the adaptation, such as ``in reverse order''.
With no consideration of skill-level cross-domain contexts, we test 27 and 24 unseen task prompts for Franka Kitchen and Meta-World, respectively. Table~\ref{exp:task-level} shows that $\oursol$ exhibits robust performance for the ``Reverse'' and ``Replace'' cases with $N$-rate decline of $7.08\%$ and $5.97\%$ in Franka Kitchen and Meta-world, respectively, compared to the single-domain reference performance (``Ref.''). In contrast, TP-GPT degrades by $9.39\%$ and $2.5\%$ in $N$-rate. This specifies the task-level adaptation capabilities of $\oursol$, specifically empowered by the semantic skill decoder based on a PLM.
\begin{table}[h]
    \small
    \centering
    {
    \begin{tabular}{c|c||c|c|c|c|c}
    \specialrule{.1em}{.05em}{.05em}
    \textbf{Env} & \textbf{Dom} & \textbf{VIMA} & \textbf{BCZ} & \textbf{GPT} & \textbf{VCP} & \textbf{SemTra} \\
    \specialrule{.1em}{.05em}{.05em}

    \multirow{3}{*}{FK} 
    & Ref. & 48.1 & 22.2 & 59.2 & 43.5 & \textbf{73.1} \\ 
    \cline{2-7}
    & Rev & 46.3 & 14.8 & 50.9 & 62.9 & \textbf{71.3} \\ 
    & Rep & 30.9 & 15.4 & 48.8 & 26.1 & \textbf{64.2} \\ 
    \specialrule{.1em}{.05em}{.05em} 

    \multirow{3}{*}{MW} 
    & Ref. & 16.6 & 15.2 & 22.2 & 20.8 & \textbf{80.5} \\ 
    \cline{2-7}
    & Rev & 5.5 & 9.7 & 8.3 & 8.3 & \textbf{76.3} \\ 
    & Rep & 8.8 & 8.8 & 13.3 & 5.5 & \textbf{80.0} \\ 
    \specialrule{.1em}{.05em}{.05em} 
    
    \end{tabular}
    }
    \caption{Task-level cross-domain performance in $N$-rate.}
    \label{exp:task-level}
\end{table} 

\subsubsection{Skill-level cross-domain performance.}
Table~\ref{exp:skill-level} shows the cross-domain adaptation performance, particularly for skill-level domain changes, in which each skill or subtask is required to adapt to the target domain. For each instructed domain context, we compute the average $N$-rate performance across different non-stationary hidden contexts. For example, ``Fast'' in the ``Domain'' column indicates domain contexts associated with stringent time limits in the environment, which can be instructed (specified) by a given task prompt; the snippets beneficial for the skill-level cross-domain context, such as ``do fast'', are given.
For each row in the table, we experiment with 36 and 32 different task prompts for Franka Kitchen and Meta-World, respectively.
$\oursol$ exhibits robust performance compared to the baselines, showing $39.35\%$ and $65.73\%$ higher in Franka Kitchen and Meta-World than the most competitive baselines TP-BCZ and TP-GPT, respectively. This is attributed to our disentanglement scheme, which effectively separates domain-invariant semantic skills from domain contexts. 
\begin{table}[h]
    \small
    \centering
    {
    \begin{tabular}{c|c||c|c|c|c|c}
    \specialrule{.1em}{.05em}{.05em}
    \textbf{Env} & \textbf{Dom} & \textbf{VIMA} & \textbf{BCZ} & \textbf{GPT} & \textbf{VCP} & \textbf{SemTra} \\
    \specialrule{.1em}{.05em}{.05em}

    \multirow{3}{*}{FK} 
    & Ref. & 48.1 & 22.2 & 59.2 & 43.5 & \textbf{73.1} \\ 
    \cline{2-7}\
    & Gust & 13.6 & 23.1 & 12.7 & 15.7 & \textbf{59.9} \\ 
    & Flurry & 7.8 & 19.4 & 4.4 & 16.6 & \textbf{61.3} \\ 
    \specialrule{.1em}{.05em}{.05em} 

    \multirow{3}{*}{MW} 
    & Ref. & 16.6 & 15.2 & 22.2 & 20.8 & \textbf{80.5} \\ 
    \cline{2-7} 
    & Slow & 1.0 & 12.5 & 9.3 & 4.1 & \textbf{72.1} \\ 
    & Fast & 3.4 & 0.0 & 11.4 & 15.2 & \textbf{80.2} \\ 
    \specialrule{.1em}{.05em}{.05em} 
    
    \end{tabular}
    }
    \caption{Skill-level cross-domain performance in $N$-rate.}
    \label{exp:skill-level}
\end{table}

\subsubsection{Semantic correspondence in multi-modal space.}
\begin{figure}[h]
    \centering
    \includegraphics[width=0.46\textwidth]{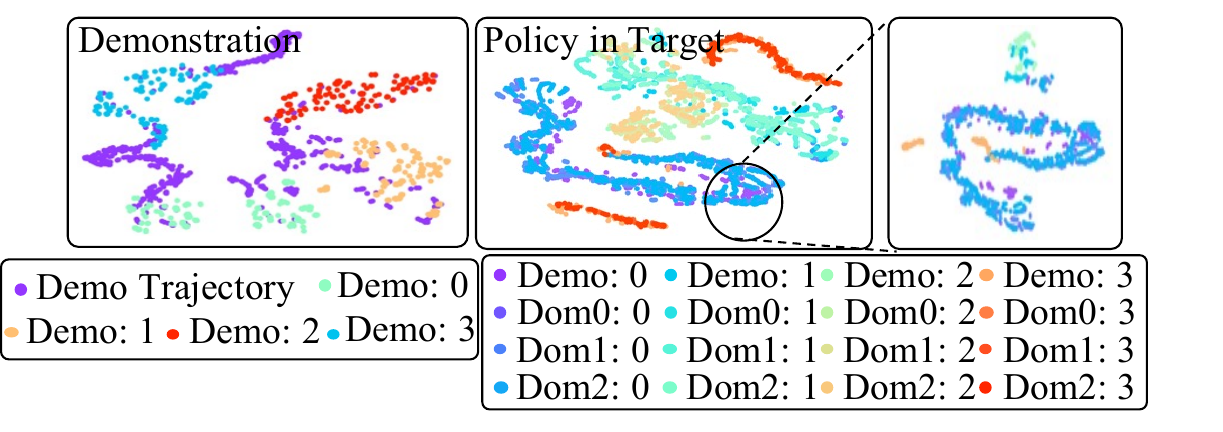}
    \caption{
    Semantic correspondence in V-CLIP space.}
    \label{fig:analysis_semantic}
\end{figure}

In Figure~\ref{fig:analysis_semantic}, we visualize a set of video demonstrations and the trajectories of the zero-shot policy in $\oursol$. 
We use the V-CLIP embedding space. A numeric index in the legends corresponds to a semantic skill. 
As observed, the demonstration trajectories (on the left side of the figure) traverse through distinct clusters of semantic skills, reflecting the alignment of the skills with the expert behavior patterns. 
The policy's trajectories (in the middle of the figure) in the target domains share a similar space with the demonstrations. 
In addition, the trajectories from different domains (``Dom''s in legends) for a single semantic skill (on the right side of the figure) tend to form distinct clusters, indicating the variations in the execution of a specific skill in different domain contexts.

\subsection{Ablation Studies}
\label{sec: ablation studies}

\subsubsection{PLMs for skill sequence generation.}
\begin{figure}[h]
    \centering
    \includegraphics[width=0.50\textwidth]{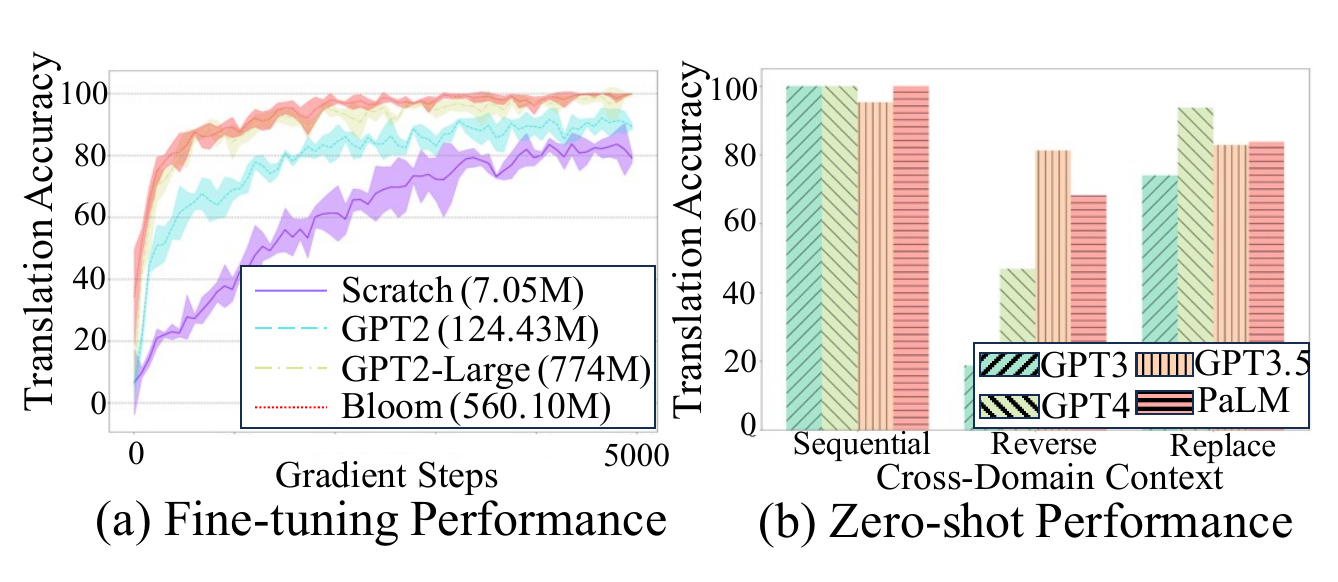}
    \caption{
    PLMs for skill sequence generator: 
    (a) PLM fine-tuning case; the x-axis denotes the gradient update steps, and the y-axis denotes the accuracy of the skill sequence generation. 
    (b) PLM zero-shot case; only an engineered prompt is adopted without any model fine-tuning.  
    }
    \label{fig:llm translator}
\end{figure}

To implement the semantic skill sequence generator $\Phi_G$ in~\eqref{sequence generator}, we test several PLMs, including relatively lightweight models such as GPT2, GPT2-large~\cite{radford-gpt2}, and Bloom~\cite{bloom}. These models are selected based on the consideration that they can be fine-tuned with the GPU system resources available in an academic setting. In Figure~\ref{fig:llm translator}, these PLMs consistently exhibit higher sample efficiency than the model trained from scratch. There is also a positive correlation between the size of the PLMs and the accuracy of skill sequence generation. This specifies that the logical reasoning abilities of PLMs are instrumental in task adaptation.

We also evaluate the suitability of advanced PLMs such as GPT3~\cite{brown_gpt3}, GPT3.5, GPT4~\cite{gpt4_report}, and PaLM~\cite{palm} for the skill sequence generator, in a \textit{zero-shot} context, without any fine-tuning. 
To this end, we engineer an augmented task prompt to contain a semantic skill set, which is presented to the PLMs along with the skill-level instruction. Our observations reveal that GPT3 and GPT3.5 encounter difficulties in generating the correct sequence, especially when reversing the execution of a skill sequence, while GPT4 and PaLM achieve relatively better performance. These results are consistent with our motivation to leverage the advancements in PLMs in enabling robust translation from intermediate instructions to semantic skill sequences. 

\subsubsection{Conditional components for behavior decoding.}
We test several representations for the conditional components to the behavior decoder in~\eqref{eqn: behavior decoder} to evaluate our skill adaptation phase. Specifically, we explore variations in the inputs to the behavior decoder instead of executable skills $\bar{z_t}$ in~\eqref{eqn: behavior decoder}:
\textbf{N-GPT} uses a given task prompt in multi-modal snippets in~\eqref{eqn: task interpretation io}; 
\textbf{L-GPT} uses a skill-level instruction in~\eqref{eqn: multi-modal skill encoder} that has been encoded in the language embedding space by the multi-modal encoders; 
\textbf{V-GPT} is the same as L-GPT but employs the video embedding space;
\textbf{T-GPT} takes the entire sequence of executable skills in~\eqref{eqn: context encoder} bypassing the sequence segmentation in~\eqref{eqn: boundary detector}. 

In Table~\ref{tbl:ablation_condition}, comparing N-GPT and $\{\textnormal{V, L}\}$-GPT, we observe that retrieving semantics from video demonstrations improves performance in $N$-rate by $16.45\%$. L-GPT outperforms V-GPT, specifying that skill embeddings in the language space are more efficient than those in the video space; this pattern was observed in previous work~\cite{bc-z}. 
T-GPT outperforms L-GPT by $12.54\%$, indicating the effectiveness of the skill sequence generation of our framework. Finally, $\oursol$ outperforms T-GPT by $35.57\%$, demonstrating that a behavior decoder conditioned with a single skill is suitable. 
These consistent results clarify the benefits of our skill adaptation based on disentangled skill semantics and target domain contexts, enabling parametric skill instantiation. 

\begin{table}[h]   
    \small
    \centering
    {
        \begin{tabular}{c|c|c|c|c|c}
        \specialrule{.1em}{.05em}{.05em}
        \textbf{Env} 
        & \multicolumn{1}{c|}{\textbf{N-GPT}} & \multicolumn{1}{c|}{\textbf{V-GPT}} & \multicolumn{1}{c|}{\textbf{L-GPT}}
        & \multicolumn{1}{c|}{\textbf{T-GPT}} & \multicolumn{1}{c}{\textbf{SemTra}}
        \\
        \specialrule{.1em}{.05em}{.05em}

        FK
        & 4.1 & 19.4 & 40.2 & 42.8 & \textbf{72.2} \\
        MW
        & 2.7 & 8.1 & 11.8 & 34.3 & \textbf{76.1} \\
        \specialrule{.1em}{.05em}{.05em}

        \end{tabular}
    }
    \caption{Conditional components for behavior decoding.}
    \label{tbl:ablation_condition}
\end{table}

\subsubsection{Sample efficiency in semantic annotations.}
As explained in Section~\ref{sec: task interpretation}, we leverage a training dataset with semantically annotated target trajectories. Recognizing the challenges associated with acquiring a large annotated dataset, we employ the video skill encoder, which was trained on a small annotated dataset, to pseudo-label on unannotated demonstrations. We evaluate the impact of the annotated dataset size on the efficacy of our framework. Table~\ref{exp: sem_ann_efficiency} shows the accuracy of the video skill encoder (in the row of Skill Pred.) and the cross-domain adaptation performance ($N$-rate) with respect to the number of annotations used. Remarkably, utilizing only 50 videos per semantic skill (approximately 0.23\% of the entire dataset) is sufficient for our framework to achieve comparable performance to the cases where annotations are provided at every timestep.

\begin{table}[h]
    \small
    \centering
    {
    \begin{tabular}{c||c|c|c|c|c|c}
    \specialrule{.1em}{.05em}{.05em}
    \textbf{\# Ann.} & \textbf{3} & \textbf{5} & \textbf{7} & \textbf{10} & \textbf{50} & \textbf{Every}\\
    \specialrule{.1em}{.05em}{.05em}
    Skill Pred. & 68.5 & 68.5 & 84.4 & 89.6 & 99.7 & 100.0\\ 
    $N$-rate & 45.8 & 46.6 & 64.7 & 64.4 & 71.1 & 72.7\\ 
    \specialrule{.1em}{.05em}{.05em} 
    \end{tabular}
    }
    \caption{Sample efficiency in semantic annotations}
    \label{exp: sem_ann_efficiency}
\end{table}

\subsection{Use Cases}

\subsubsection{Cognitive robot manipulator.}
We verify the applicability of $\oursol$ with cognitive robot manipulation scenarios using the RLBench robotic manipulation simulator~\cite{rlbench}, in which a task prompt is presented as \textit{abstract user commands} in natural language (contrast with our task prompts containing video demonstrations.) Specifically, we experiment with three different command styles, similar to the applications of SayCan~\cite{saycan}.
A \textbf{Summarized sentence} command offers an abstract instruction for conveying long-horizon tasks.   
An \textbf{Abstract verbs} command encompasses synonymous verbs to convey desired actions. 
An \textbf{Embodiment} command narrates the current environment conditions and the constraints of the robot manipulator.
Table~\ref{exp: cognitive robot} demonstrates the efficacy of $\oursol$ in processing these abstract commands, specifying the comprehensive task prompting structure as well as the ability to harness the logical reasoning capabilities of PLMs. Furthermore, the two-phase adaptation facilitates modular approaches for different levels of policy adaptations using PLMs. It also supports the flexible integration of future, more advanced PLMs.
\begin{table}[h]
    \small
    \centering
    {
        \begin{tabular}{c||cc|ccc}
        \specialrule{.1em}{.05em}{.05em}
        \multirow{2}{*}{\textbf{}} 
        & \multicolumn{2}{c|}{\textbf{SayCan}} & \multicolumn{3}{c}{\textbf{SemTra}} \\
        \cline{2-3} \cline{4-6}
        & \multicolumn{1}{c|}{\textbf{GPT-3}} & \multicolumn{1}{c|}{\textbf{PaLM}} & \multicolumn{1}{c|}{\textbf{GPT-3.5}} & \multicolumn{1}{c|}{\textbf{GPT-4}} & \multicolumn{1}{c}{\textbf{PaLM}} \\
        \specialrule{.1em}{.05em}{.05em}
        Summ. & \multicolumn{1}{c|}{76.6} & \multicolumn{1}{c|}{100.0} & \multicolumn{1}{c|}{53.3} & \multicolumn{1}{c|}{100.0}  & \multicolumn{1}{c}{100.0} \\ 
        Abst. & \multicolumn{1}{c|}{60.0}  & \multicolumn{1}{c|}{100.0} & \multicolumn{1}{c|}{73.3} & \multicolumn{1}{c|}{100.0} & \multicolumn{1}{c}{100.0} \\ 
        Emb. & \multicolumn{1}{c|}{86.6} & \multicolumn{1}{c|}{100.0} & \multicolumn{1}{c|}{60.0}  & \multicolumn{1}{c|}{100.0} & \multicolumn{1}{c}{100.0}  \\ 
        \specialrule{.1em}{.05em}{.05em}
        \end{tabular}
    }
    \caption{Performance in cognitive robot manipulations.}
    \label{exp: cognitive robot}
\end{table}
\color{black}

\subsubsection{Autonomous driving.}
\begin{figure}[h]
    \centering
    \includegraphics[width=0.48\textwidth]{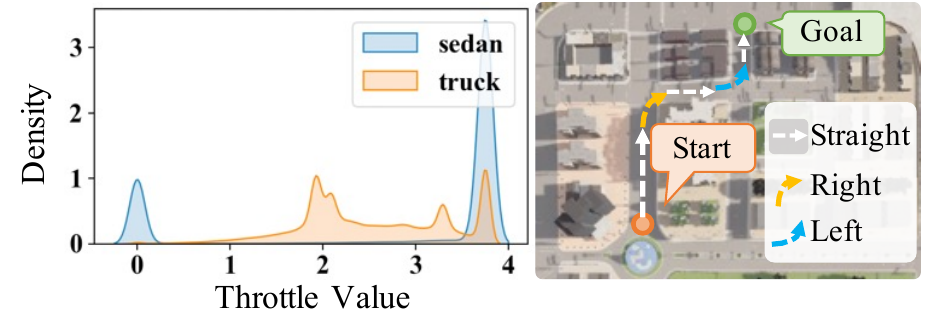}
    \caption{
    The left graph represents the difference in expert action distribution based on  vehicle configurations. 
    The right figure shows a semantic skill sequence to reach the goal.
    }
    \label{fig:carla}
\end{figure}

With the autonomous driving simulator CARLA~\cite{carla}, we verify the skill adaptation ability across different vehicle configurations (agent embodiments) such as sedans and trucks. The driving agent must calibrate steering, throttle, and brake responses according to the specific vehicle configuration it manages. Its goal is to reach the destination in the shortest path as depicted in Figure~\ref{fig:carla}, where skills are semantically defined as driving straight, turning left, and turning right.
We calculate the normalized return based on the rewards yielded by a rule-based policy. As shown in Table~\ref{exp:carla}, $\oursol$ outperforms the most competitive baseline VCP by $1.94 \sim 5.00\%$.

\begin{table}[h]
    \small
    \centering
    {
    \begin{tabular}{c||c|c|c|c|c}
    \specialrule{.1em}{.05em}{.05em}
    \textbf{Vehicle type} & \textbf{VIMA} & \textbf{BCZ} & \textbf{GPT} & \textbf{VCP} & \textbf{SemTra} \\
    \specialrule{.1em}{.05em}{.05em}

    Sedan & 79.4 & 74.7 & 62.7 & 78.3 & \textbf{83.3} \\ 
    Truck & 33.4 & 68.6 & 65.1 & 70.5 & \textbf{72.4} \\ 
    \specialrule{.1em}{.05em}{.05em} 
    \end{tabular}
    }
    \caption{
    Performance in autonomous driving.
    }
    \label{exp:carla}
\end{table}

\section{Related Works}
\subsubsection{Cross-domain policy adaptation.}
Recently, there has been a surge in cross-domain policy adaptation techniques exploring task specifications~\cite{goyal:retail, pertsch:star}. 
A relevant subset of this trend is cross-domain one-shot imitation, which exploits demonstrations from the source domain to tackle tasks in different target domains~\cite{mandi_mosaic, kuno_dail, liu_imio, daml, gupta_inv_feat}. 
For instance, \cite{mandi_mosaic} introduced a contrastive learning-based video encoder to bridge visual domain gaps between demonstrations and target environments. \cite{kuno_dail} employed the generative adversarial learning to handle embodiment and dynamics differences, while \cite{goyal:retail} delved into task-level domain disparities, directly transferring source expert trajectories to target actions using a Transformer-based policy.
Our work is in the same vein, but it distinguishes itself by accommodating a unified multi-modal specification for task prompting. We also concentrate on long-horizon tasks, incorporating the knowledge of PLMs into our framework to explore the compositionality and parametric execution capability of semantic skills.

\subsubsection{Semantic skills in RL.}
In the RL literature, several studies introduced policy learning and adaptation methods, particularly investigating semantically interpretable skills (semantic skills). 
Recently, Saycan~\cite{saycan} employed a PLM to plan semantic skills from abstract linguistic instructions. Utilizing an offline RL dataset collected from an environment with domain-specific rewards, this novel approach learns to estimate the value of each semantic skill at a specific scene, effectively opting to execute the most valuable skill. 
Furthermore, \cite{pertsch:star} exploited human demonstrations with semantic annotations via state distribution matching to expedite RL training.
However, our emphasis lies on their ability to adapt in a zero-shot manner across domains by developing a novel two-phase adaptation process.

\subsubsection{Multi-modality in robotic manipulation.}
In recent studies on robotic manipulation, several approaches leveraging large-scale pretrained multi-modal models have been introduced, showing promise in enhanced policy generalization by adeptly integrating both visual and language instructions~\cite{guhur2023instruction, shridhar2022cliport, liu2022instruction}. \cite{bc-z} proposed a multi-modal imitation learning framework, which trains the vision encoder by maximizing the similarity of video embeddings and the PLMs embedding vectors of video descriptions. \cite{jiang2023vima} proposed a unified robot-manipulation framework capable of integrating interleaved video-language inputs by encoding them with pretrained vision and language models.
To the best of our knowledge, $\oursol$ is the first to consider a unified multi-modal task prompt and incorporate the knowledge of PLMs and parametric semantic skills into cross-domain policy adaptation procedures.

\section{Conclusion}
We presented the semantic skill translator $\oursol$, a framework that leverages the knowledge of PLMs to achieve zero-shot policy adaptation across different domains. 
In the framework, we employ a two-tiered adaptation process in which a sequence of multi-modal interleaved snippets serves as a task prompt, facilitating task-level semantic skill generation and skill-level refinement tailored to the target domain contexts.
In our future work, we will explore the versatility of PLMs more under intricate cross-domain conditions that require dynamic adjustments of a skill sequence based on real-time observations.  
%

\section*{Acknowledgements}
We would like to thank anonymous reviewers for their valuable comments and suggestions. 
This work was supported by 
Institute of Information \& communications Technology Planning \& Evaluation (IITP) grant funded by the Korea government (MSIT) (No.
2022-0-01045, 
2022-0-00043,
2019-0-00421, 
2020-0-01821) 
 and by the National Research Foundation of Korea (NRF) grant funded by the MSIT 
(No. RS-2023-00213118). 

\bibliography{ref.bib}

\appendix

\lstset{
  basicstyle=\ttfamily\footnotesize,  
  breaklines=true,
  numbers=none,
  xleftmargin=0pt,
}

\lstdefinestyle{mypy}{
    language=Python,
    basicstyle=\ttfamily\small,
    showstringspaces=false,
    breaklines=true,
    frame=tb, 
    framerule=0.5pt,
    rulecolor=\color{black},
    numbers=none, 
    aboveskip=1em, 
    belowskip=1em, 
    columns=flexible,
    captionpos=t 
}

\renewcommand{\theequation}{A.\arabic{equation}}
\renewcommand{\thefigure}{A.\arabic{figure}}
\renewcommand{\thetable}{A.\arabic{table}}
\setcounter{secnumdepth}{2}
\setcounter{equation}{0}

\section{Implementation Details}

\subsection{SemTra Implementation}
Our model is implemented using Python v3.8.16 and automatic gradient framework Jax v0.4.4. 
We train the models on a system with an NVIDIA RTX 4090 GPU.
$\oursol$ consists of six parameterized learnable models: multi-modal skill encoder $\Phi_E$, semantic skill sequence generator $\Phi_G$, skill boundary detector $\Phi_B$, context encoder $\Phi^{(g)}_C$, online context encoder $\Phi^{(o)}_C$, and behavior decoder $\pi$. We will now provide detailed explanations of their implementations.

\subsubsection{Multi-modal skill encoder.}
To implement the video skill encoder $\Psi_V$, which maps a video snippet $x = (v_{0:T})$ to skill-level language tokens, we utilize the Video-CLIP (V-CLIP)~\cite{v-clip} which consists of a video encoder $\Psi_V$ and a language encoder $\Psi_L$. The implementation of this video skill encoder is based on an open-source project by~\cite{ott2019fairseq}. The prompt $\theta_p$ is a set of learnable parameters, which is concatenated with the text to form a comprehensive input vector for the language encoder $\Psi_L$ in~\eqref{eqn: video skill encoder}.

For mapping a sensor snippet $x = (s_{t-H': t}, a_{t-H': t})$ to a skill-level language token, we utilize a Transformer-based classifier~\cite{attention_is_all_you_need} $\Psi_S$, which predicts the corresponding semantic annotation $l_t$.
The hyperparameter settings for the multi-modal skill encoder $\Phi_E$ are summarized in Table~\ref{appendix: hyperparameter: multi-modal skill encoder}.
\begin{table}[h]
\centering
        \resizebox{7cm}{!}
        {
            \begin{tabular}{c|c}
            \hline
                \textbf{Hyperparameter} & \textbf{Value} \\
                \hline
                Size of $\theta_p$ & 16 vector in 768-dim \\
                Batch size for $\theta_p$ & 1 \\
                Learning rate for $\theta_p$ & 1e-4 \\

                The number of layer for $\Psi_S$ & 4 \\
                The number of head for $\Psi_S$ & 4 \\
                Activation fn for $\Psi_S$ & Relu \\
                Dropout for $\Psi_S$ & 0.1 \\
                Hidden dimension for $\Psi_S$ & 256 \\
                Batch size for $\Psi_S$ & 256 \\
                Learning rate for $\Psi_S$ & 5e-4 \\
            \hline
    	\end{tabular}
        }
        \caption{Hyperparameters for the multi-modal skill encoder.}
        \label{appendix: hyperparameter: multi-modal skill encoder}
\end{table}

\subsubsection{Semantic skill sequence generator.}
To train the semantic skill sequence generator $\Phi_G$, we employ a relatively small-sized pretrained language model GPT2~\cite{radford-gpt2} with 124.43 million parameters, obtained from the open-source Huggingface library~\cite{huggingface_gpt2}. Other pretrained models used in Figure~\ref{fig:llm translator} of the main manuscript were also obtained from Huggingface's models~\cite{huggingface_gpt2, huggingface_bloom}. For the model from scratch $\Phi^{(scr)}_G$ in Figure~\ref{fig:llm translator}, we train an encoder-decoder Transformer having 7.05 million parameters from scratch, using the Bert~\cite{bert} token embedding.
The hyperparameter settings for the semantic skill sequence generator $\Phi_G^{(scr)}$ are summarized in Table~\ref{appendix: hyperparameter: skill sequence generator}.
\begin{table}[h]
\centering
        \resizebox{7cm}{!}
        {
            \begin{tabular}{c|c}
            \hline
                \textbf{Hyperparameter} & \textbf{Value} \\
                \hline
                Batch size for $\Phi^{(GPT)}_G$ & 3 \\
                Learning rate for $\Phi^{(GPT)}_G$ & 1e-5 \\
                
                Batch size for $\Phi^{(scr)}_G$ & 3 \\
                Learning rate for $\Phi^{(scr)}_G$ & 1e-4 \\
                The number of layer of the encoder for $\Phi^{(scr)}_G$ & 1 \\
                The number of head of the encoder for $\Phi^{(scr)}_G$ & 8 \\
                The number of layer of the decoder for $\Phi^{(scr)}_G$ & 1 \\
                The number of head of the decoder for $\Phi^{(scr)}_G$ & 4 \\
                Feed-forward network dim. for $\Phi^{(scr)}_G$ & 32 \\
                Activation fn for $\Phi^{(scr)}_G$ & Relu \\
                Dropout for $\Phi^{(scr)}_G$ & 0.15 \\
            \hline
    	\end{tabular}
        }
        \caption{Hyperparameters for the skill sequence generator.}
        \label{appendix: hyperparameter: skill sequence generator}
\end{table}

\subsubsection{Skill boundary detector.}
To train the skill boundary detector $\Phi_B$, we utilize pseudo-labeling by assigning each timestep of the expert trajectory $\tau^{(tgt)}$ with the predicted semantic skill $z_t = \Psi_V(v_{t-H: t})$, using the video skill encoder $\Psi_V$. As discussed in Section~\ref{sec: ablation studies} of the main manuscript, if semantic labels already exist for all timesteps, we directly use them as $z_t$. Since the number of times the semantic skill changes in a target trajectory is relatively low compared to the total length of timesteps, we observe that this results in a low proportion of positive labels for the skill boundary detector to predict. To address this, we extend the positive labels around the timesteps where the actual semantic skill changes by $\pm J$. The hyperparameter settings for the skill boundary detector $\Phi_B$ are summarized in Table~\ref{appendix: hyperparameter: skill boundary detector}.
\begin{table}[h]
        \centering
        \resizebox{5cm}{!}
        {
            \begin{tabular}{c|c}
            \hline
                \textbf{Hyperparameter} & \textbf{Value} \\
                \hline
                Batch size for $\Phi^{(g)}_G$ & 3 \\
                Learning rate for $\Phi^{(g)}_G$ & 1e-5 \\
            \hline
    	\end{tabular}
        }
        \caption{Hyperparameters for the skill boundary detector.}
        \label{appendix: hyperparameter: skill boundary detector}
\end{table}

\subsubsection{Context encoder.}
To train the context encoder $\Phi^{(g)}_C$, we fine-tune a pretrained GPT2 model~\cite{huggingface_gpt2} similar to the semantic skill sequence generator. The context encoder takes natural language input, which consists of a skill-level language instruction derived from the task prompt and a semantic skill sequence connected with conjunctions. The target output is the executable skill sequence for the target trajectory. Each executable skill is represented as a tuple in natural language form, consisting of (skill semantic, domain context, magnitude), e.g., (``Open door'', ``Speed'', ``Slow''). The procedure of creating the target trajectory and the corresponding task prompt for each trajectory is described in detail in Appendix Section~\ref{appendix: sec: environments and datasets}. The hyperparameter settings for the context encoder $\Phi_C$ are summarized in Table~\ref{appendix: hyperparameter: context encoder}.
\begin{table}[h]
\centering
        \resizebox{7cm}{!}
        {
            \begin{tabular}{c|c}
            \hline
                \textbf{Hyperparameter} & \textbf{Value} \\
                \hline
                Network architecture for $\Phi_B$ & 4 FC with 128 units \\
                Batch size for $\Phi_B$ & 256 \\
                Learning rate for $\Phi_B$ & 1e-4 \\
                Activation fn for $\Phi_B$ & Leaky Relu \\
                Dropout for $\Phi_B$ & 0.0 \\
                Positive label augmentation $J$ & 4 \\
            \hline
    	\end{tabular}
        }
        \caption{Hyperparameters for the context encoder}
        \label{appendix: hyperparameter: context encoder}
\end{table}

\subsubsection{Online context encoder.}
To train the online context encoder $\Phi^{(o)}_C$, we adopt a contrastive learning-based dynamics encoder, as proposed in~\cite{onis, ood_imitation}. Specifically, this model takes an expert sub-trajectory $\tau = (s_{t-H':t}, a_{t-H': t-1})$ as input and encodes the dynamics parameter $h_t$ of the environment. This is quantized into a discrete value using vector quantization~\cite{oord_quantization}. For contrastive learning, two different sub-trajectories are considered as positive pairs if they are sampled from the same trajectory. The online context encoder $\Phi^{(o)}_C$ is jointly trained with the inverse dynamics decoder $\psi_{dec}$, which takes the output $h_t$ of $\Phi^{(o)}_C$ as input and reconstructs the sub-trajectory. The learning process is controlled by the coefficient $\alpha$ of the contrastive loss. The hyperparameter settings for the online context encoder $\Phi^{(o)}_C$ are summarized in Table~\ref{appendix: hyperparameter: online context encoder}.
\begin{table}[h]
\centering
        \resizebox{8cm}{!}
        {
            \begin{tabular}{c|c}
            \hline
                \textbf{Hyperparameter} & \textbf{Value} \\
                \hline
                Network architecture for $\Phi^{(o)}_C$ & LSTM with 32 units \\
                Batch size for $\Phi^{(o)}_C$ & 256 \\
                Learning rate for $\Phi^{(o)}_C$ & 1e-4 \\
                Activation function for $\Phi^{(o)}_C$ & Relu \\
                Sub-trajectory length $H'$ & 3 \\
                Coefficient $\alpha$ for Meta-World & 1.0 \\
                Coefficient $\alpha$ for Franka Kitchen & 0.1 \\
                Dimension of codebook for $\Phi^{(o)}_C$ & 16 \\
                The number of codebooks for $\Phi^{(o)}_C$ & 50 \\

                Network architecture for $\psi_{dec}$ & 2 FC with 32 units \\
                Batch size for $\psi_{dec}$ & 256 \\
                Learning rate for $\psi_{dec}$ & 1e-4 \\
                Activation function for $\psi_{dec}$ & Relu \\
            \hline
    	\end{tabular}
        }
        \caption{Hyperparameters for the online context encoder}
        \label{appendix: hyperparameter: online context encoder}
\end{table}

\subsubsection{Behavior decoder.}
To train the behavior decoder $\pi$, we utilize a multi-layer perceptron (MLP) with layer normalization~\cite{lei_layer_normalization}. The hyperparameter settings for the behavior decoder $\pi$ are summarized in Table~\ref{appendix: hyperparameter: behavior decoder}.
\begin{table}[h]
\centering
        \resizebox{\columnwidth}{!}
        {
            \begin{tabular}{c|c}
            \hline
                \textbf{Hyperparameter} & \textbf{Value} \\
                \hline
                Network architecture of $\pi$ for Meta-World & 7 FC with 128 units \\
                Network architecture of $\pi$ for Franka kitchen & 4 FC with 128 units \\
                Batch size for $\pi$ & 256 \\
                Learning rate for $\pi$ & 1e-4 \\
                Activation function of $\pi$ for Meta-World & Relu \\
                Activation function of $\pi$ for Franka kitchen & Leaky Relu \\
            \hline
    	\end{tabular}
        }
        \caption{Hyperparameters for the behavior decoder}
        \label{appendix: hyperparameter: behavior decoder}
\end{table}

\subsection{Baseline Implementation}
\subsubsection{VIMA.}
VIMA is a framework that accomplishes various robotics tasks, such as one-shot imitation or rearrangement, through interleaved multi-modal prompting of video and language. We adopt the same tokenizing technique as presented in the original paper to address multi-modal snippets. Specifically, the frames of video snippets are embedded using a pretrained vision Transformer~\cite{vit}. Text snippets are embedded using a pretrained language model T5, with the last 2 layers fine-tuned through behavior cloning loss. As the original paper does not cover sensor snippets, we additionally train an MLP $\Psi_S$ with the same output dimension as the embeddings of other snippets, using behavior cloning loss. The embeddings of each snippet are concatenated with the current state $s_t$ and used as input to a cross-attention-based Transformer $\pi$. The hyperparameter settings for VIMA are summarized in Table~\ref{appendix: hyperparameter: VIMA}.
\begin{table}[h]
\centering
        \resizebox{7cm}{!}
        {
            \begin{tabular}{c|c}
            \hline
                \textbf{Hyperparameter} & \textbf{Value} \\
                \hline
                Batch size & 256 \\
                Learning rate & 1e-4 \\
                Network architecture for $\Psi_S$ & 1 FC with 768 units \\

                Learning rate for $\pi$ & 1e-4 \\
                The number of layers for $\pi$ & 3 \\
                The number of heads for $\pi$ & 4 \\
                Subsequence length & 20 \\
            \hline
    	\end{tabular}
        }
        \caption{Hyperparameters for VIMA}
        \label{appendix: hyperparameter: VIMA}
\end{table}

\subsubsection{TP-BCZ.}
BCZ~\cite{bc-z} is the state-of-the-art in multi-modal imitation learning framework, which employs contrastive learning to increase similarity in the multi-modal space between videos and their corresponding instructions or captions. To adapt this idea to the cross-domain context, we implement TP-BCZ, which trains an additional regression network $\psi_{g}$ which performs contrastive learning to maximize the similarity between the embeddings of task prompt $\TP$ and the embeddings of language instruction of the corresponding target trajectory. The embedding method for each snippet in the task prompt is the same as in $\oursol$, and we construct the language instruction of the target trajectory by concatenating the semantic skill sequence with the conjunction ``and''. The policy $\pi$ is conditioned on the embeddings of task prompt $\TP$.
The hyperparameter settings for TP-BCZ are summarized in Table~\ref{appendix: hyperparameter: TP-BCZ}.
\begin{table}[h]
\centering
        \resizebox{7cm}{!}
        {
            \begin{tabular}{c|c}
            \hline
                \textbf{Hyperparameter} & \textbf{Value} \\
                \hline
                Batch size & 256 \\
                Network architecture for $\psi_g$ & 3 FC with 256 units \\
                Learning rate for $\psi_g$ & 1e-4 \\

                Network architecture for $\pi$ & 4 FC with 128 units \\
                Learning rate for $\pi$ & 1e-4 \\
            \hline
    	\end{tabular}
        }
        \caption{Hyperparameters for TP-BCZ}
        \label{appendix: hyperparameter: TP-BCZ}
\end{table}

\subsubsection{TP-GPT.}
TP-GPT is implemented based on a Decision Transformer~\cite{chen_decision_transformer}, a model tailored for imitation learning. To handle multi-modal snippets, we utilize pretrained large models, where each frame of the video snippet is embedded using the image encoder of CLIP~\cite{clip}, and the text snippet is embedded using the language model Bert~\cite{bert}. For processing the sensor snippet, we employ an MLP $\Psi_S$ while ensuring that it has the same dimension of output as the embeddings of other snippets. The embeddings of each snippet are concatenated with the current state $s_t$ and input into a self-attention-based Transformer $\pi$. The hyperparameter settings for TP-GPT are summarized in Table~\ref{appendix: hyperparameter: TP-GPT}.
\begin{table}[h]
\centering
        \resizebox{8cm}{!}
        {
            \begin{tabular}{c|c}
            \hline
                \textbf{Hyperparameter} & \textbf{Value} \\
                \hline
                Batch size & 256 \\
                Learning rate & 1e-4 \\
                Network architecture for $\Psi_S$ & 1 FC with 768 units \\

                Learning rate for $\pi$ & 1e-4 \\
                The number of layers for $\pi$ & 3 \\
                The number of heads for $\pi$ & 4 \\
                Subsequence length & 20 \\
            \hline
    	\end{tabular}
        }
        \caption{Hyperparameters for TP-GPT}
        \label{appendix: hyperparameter: TP-GPT}
\end{table}

\subsubsection{VCP.}
We implement VCP (Video-Conditioned Policy) based on video-conditioned policies~\cite{elliot_vip, Yulin_updp}. This model trains a video embedding generator $\psi_g$, which takes the embedding of the task prompt $\TP$ as input and generates the video embedding of the target trajectory. The generated video embedding, along with the current timestep state $s_t$, is then presented to the policy $\pi$ to predict the action. For cross-domain evaluation, we explicitly employ snippet embeddings that hold cross-domain guidance for task prompting in addition to video generation and state. Similar to TP-BCZ, each snippet of the multi-modal task prompt $\TP$ is embedded using a set of pretrained large models. The hyperparameter settings for VCP are provided in Table~\ref{appendix: hyperparameter: VCP}.
\begin{table}[h]
\centering
        \resizebox{8cm}{!}
        {
            \begin{tabular}{c|c}
            \hline
                \textbf{Hyperparameter} & \textbf{Value} \\
                \hline
                Batch size & 256 \\
                Network architecture for $\psi_g$ & 3 FC with (256, 128, 256) \\
                Learning rate for $\psi_g$ & 1e-5 \\
                
                Network architecture for $\pi$ & 4 FC with 128 units \\
                Learning rate for $\pi$ & 1e-4 \\
            \hline
    	\end{tabular}
        }
        \caption{Hyperparameters for VCP}
        \label{appendix: hyperparameter: VCP}
\end{table}

\section{Environments and Datasets}
\begin{figure}[ht] 
  \begin{minipage}[b]{0.5\linewidth}
    \centering
    \includegraphics[width=.95\linewidth]{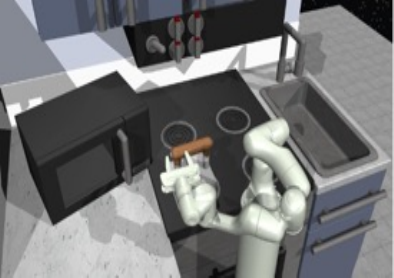} 
    \caption{Franka kitchen} 
  \end{minipage}
  \begin{minipage}[b]{0.5\linewidth}
    \centering
    \includegraphics[width=.95\linewidth]{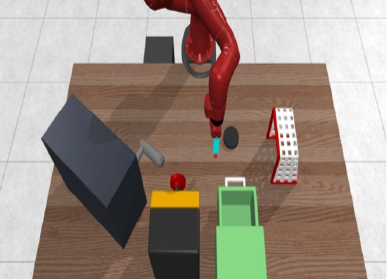} 
    \caption{Meta-World} 
  \end{minipage} 
  \begin{minipage}[b]{0.5\linewidth}
    \centering
    \includegraphics[width=.95\linewidth]{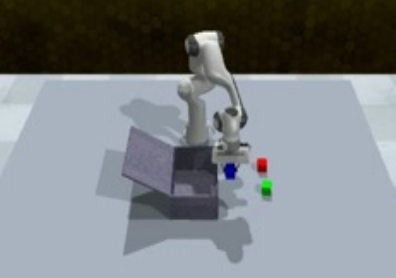} 
    \caption{RLBench} 
  \end{minipage}
  \begin{minipage}[b]{0.5\linewidth}
    \centering
    \includegraphics[width=.95\linewidth]{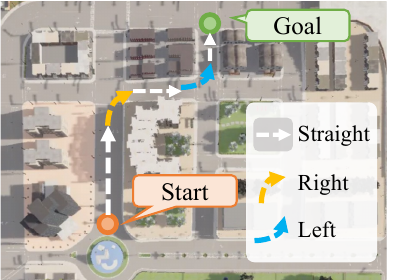} 
    \caption{CARLA} 
  \end{minipage}
  \caption{Evaluation environments}
  \label{appendix: fig: eval_environments} 
\end{figure}

We test our framework in various environments, including Franka Kitchen~\cite{d4rl}, Meta-World~\cite{metaworld}, RLBench~\cite{rlbench}, and CARLA~\cite{carla}, as described in Figure~\ref{appendix: fig: eval_environments}. In this section, we provide detailed descriptions of each environment and training dataset collection.

\label{appendix: sec: environments and datasets}

\subsection{Task prompts}
To generate training datasets, we execute a series of semantic skills across several different domain contexts. The continuity of each trajectory is maintained by setting the final state of one semantic skill as the starting point for the subsequent one. For each trajectory involving both the semantic skill sequence and the skill-level domain context, we then compute a plausible task prompt, which might have been given, based on predefined language templates. 
It is worth noting that this linkage between the task prompt and the target trajectory is a straightforward linguistic process that operates independently of trajectory generation. 
Here, we provide a typical example of a task prompt in Listing 1, and specific examples used in each environment are detailed in Sections B.2 and B.3.
Each task prompt includes placeholders \lstinline[language=Python]{SRC_TASK}, \lstinline[language=Python]{SEQ_REQ}, \lstinline[language=Python]{CONSTRAINED_SK}, and \lstinline[language=Python]{MAGNITUDE}.
\begin{itemize}
    \item \lstinline[language=Python]{SRC_TASK} contains information specifying the source task in a multi-modal form (e.g., sensor data, video demonstration performing ``Open microwave'' and then ``Light a kitchen'', or text itself).
    \item \lstinline[language=Python]{SEQ_REQ} forms the task-level cross-domain contexts. For example, if ``Replace lighting a kitchen with open a cabinet'' is provided in this placeholder \lstinline[language=Python]{SEQ_REQ}, then, combined with the example of \lstinline[language=Python]{SRC_TASK}, the agent needs to perform ``Open microwave'' and then ``Open a cabinet.''
    \item \lstinline[language=Python]{CONSTRAINED_SK} represents a specific skill or ``all skills'' to which the \lstinline[language=Python]{DEGREE} will be applied.
    \item \lstinline[language=Python]{MAGNITUDE}: Specifies the degree related to the considered domain context (e.g., ``flurry'' for the domain context wind, and ``fast'' for the domain context speed).
\end{itemize}


\begin{lstlisting}[style=mypy, caption= Task prompt examples for Franka Kitchen and Meta-World., label=listing: task prompt examples]
Ex1. "Do {SRC_TASK}, under {SEQ_REQ}. Furthermore, for {CONSTRAINED_SK} the speed {MAGNITUDE} is applied."
Ex2. "When undertaking {SRC_TASK}, ensure {SEQ_REQ} and adapt the speed {MAGNITUDE} for {CONSTRAINED_SK}."
Ex3. "While carrying out {SRC_TASK}, comply with {SEQ_REQ} and speed modifications {MAGNITUDE} to {CONSTRAINED_SK}."
Ex4. "{SRC_TASK} is performed with {SEQ_REQ} and speed {MAGNITUDE} during {CONSTRAINED_SK}."
Ex5. "During {SRC_TASK}, {SEQ_REQ} is maintained as the wind {MAGNITUDE} blows."
Ex6. "While performing {SRC_TASK} and following {SEQ_REQ}, the wind {MAGNITUDE}."
Ex7. "The wind {MAGNITUDE} blows during {SRC_TASK}, while adhering to {SEQ_REQ}."
Ex8. "{SEQ_REQ} is upheld while {SRC_TASK} and the wind {MAGNITUDE} blows."
\end{lstlisting}

\subsection{Franka Kitchen}
In this environment, a 7-DoF robot arm needs to perform sequential subtasks, such as ``open microwave'' and ``slide cabinet''.
Each trajectory in the dataset has a length of 200-500 timesteps, and the timestep limit for the target environment is set to 280. 
The domain context in this environment is represented by the wind blowing in the environment, i.e., ``breeze'', ``gust'', and ``flurry''. 

To obtain trajectories executed in such domain contexts, we use the offline dataset provided by the offline-RL benchmark D4RL~\cite{d4rl}. 
The first three axes of actions in the dataset, which represent the positions of the robot arm's grippers, are modified by adding 3 different constants $c$, similar to the approach in~\cite{onis}. The value of $c$ is set to 0.05 for ``breeze,'' 0.2 for ``gust,'' and 0.4 for ``flurry''. 
This results in different action distributions according to the instructed cross-domain contexts.
Additionally, 4 different parameters $w \in \{-0.6, -0.3, 0.3, 0.6\}$ are added to the fourth axis. 
This generates a total of 6,612 trajectories (by $551 \times 3 \times 4$). 
As the original dataset does not include semantic annotations, we manually mapped semantic annotations to each timestep by judging the video between two timesteps where positive rewards occurred (i.e., the start and completion points of a subtask).

\begin{figure}[h]
    \centering
    \includegraphics[width=0.49\textwidth]{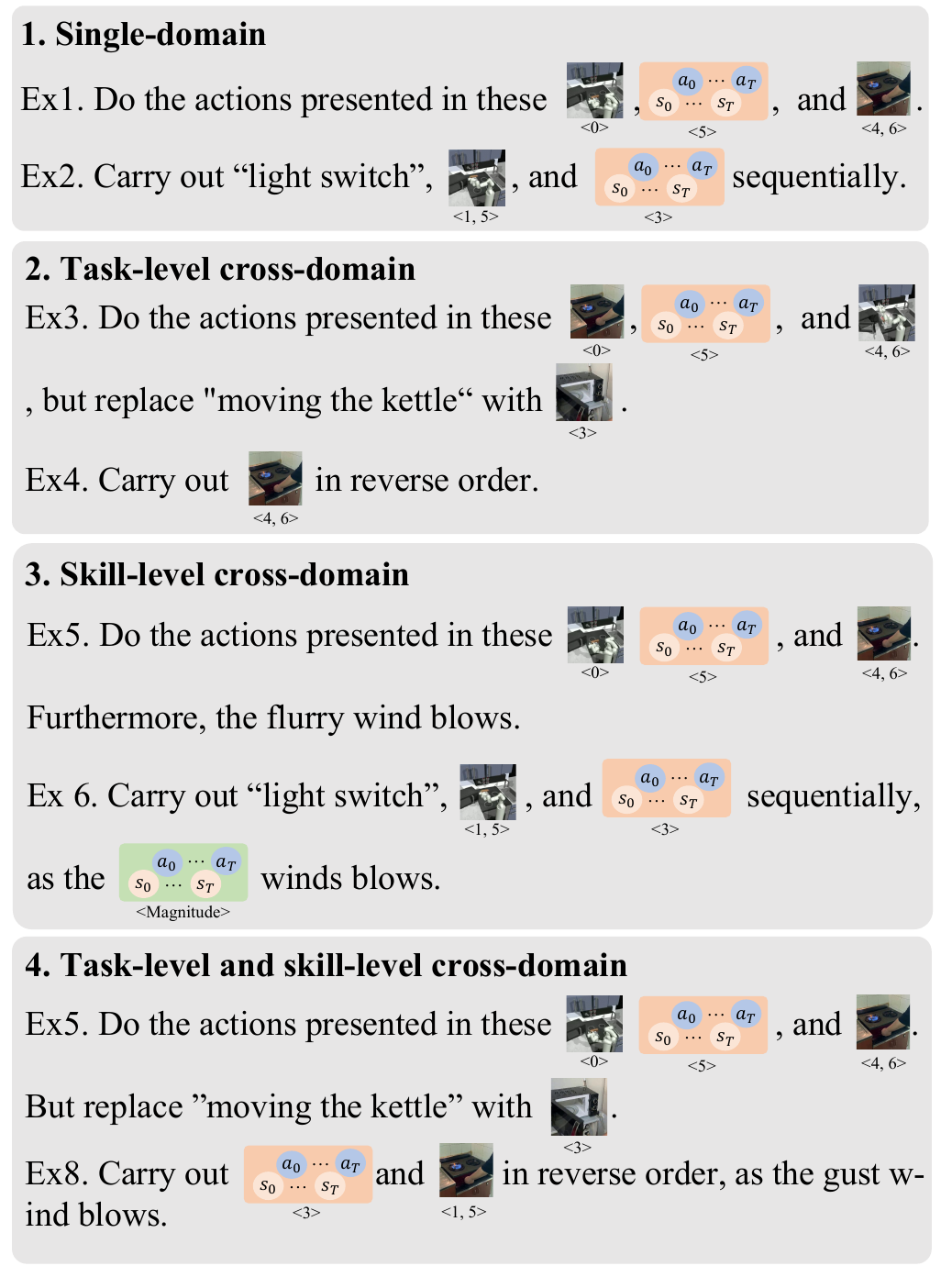}
    \caption{
    Examples of multi-modal task prompts for Franka Kitchen. The prompt comprises (video) demonstration, text, and sensor data. A numeric index $n$ in the figure corresponds to a semantic skill (e.g., 0 corresponds to ``open microwave''). Certain snippets contain cross-domain beneficial information, such as the magnitude of wind blowing in the target environment.
    }
    \label{appendix: fig: kitchen mm task prompts}
\end{figure}

During the evaluation, we introduce time-varying parameters into the fourth axis of agents' actions to simulate a non-stationary environment as explained in Section~\ref{appendix: sec: complete results}. Examples of task prompts used in Franka Kitchen are in Listing~\ref{listing: kitchen prompts} and Figure~\ref{appendix: fig: kitchen mm task prompts}.

\begin{lstlisting}[style=mypy, caption= Task prompt examples for Franka Kitchen., label=listing: kitchen prompts]
# Single-domain
Ex1. "Do the actions presented in this {SRC_TASK}."   
Ex2. "Carry out {SRC_TASK} sequentially."
# Task-level cross-domain
Ex3. "Do the actions presented in this {SRC_TASK}, but replace moving the kettle with turning on the light."  
Ex4. "Carry out {SRC_TASK} in reverse order."
# Skill-level cross-domain
Ex5. "Do the actions presented in this {SRC_TASK}. The flurry wind blows."
Ex6. "Carry out {SRC_TASK} sequentially, as the gust wind blows."
# Task-level and Skill-level cross-domain
Ex7. "Do the actions presented in this {SRC_TASK}. But replace moving the kettle with turning on the light as the blurry blows." 
Ex8. "Carry out {SRC_TASK} in reverse order, as the gust wind blows."
\end{lstlisting}

\subsection{Meta-World}
In this environment, a 3-DoF robot arm needs to perform sequential subtasks, such as ``push button'' and ``close drawer''. 
To create a multi-stage setting similar to the Franka Kitchen environment, we utilize a multi-stage Meta-World environment~\cite{onis}. 
In this environment, we use 16 training tasks, and for each task, we generate 15 episodes. 

The domain context in this environment includes the execution speed of the skill and the external wind blowing. To control the speed of semantic skill execution in three degrees (Slow, Normal, and Fast), we use a rule-based expert policy. Depending on the speeds, each trajectory in the dataset has a length of 200-900 timesteps, and the timestep limit for the target environment is set to 1,000. 
For the domain context of wind, similar to the process used in the Franka Kitchen environment, we construct three different domain contexts.
Additionally, 4 different parameters $w \in \{-0.6, -0.3, 0.3, 0.6\}$ are added to the third axis. 

As a result, this setup comprises 5,760 trajectories (by $16 \times 15 \times 2 \times 3 \times 4)$. 
We compute the number of timesteps required for executing each semantic skill in each domain context using the collected offline dataset, as shown in Table~\ref{appendix: tbl: metaworld speed}. 
During evaluations, if a task prompt includes a domain context related to the speed requirement, we determine the success or failure of the agent by calculating a confidence interval of 70\% based on the previously computed number of timesteps.

\begin{figure}[h]
    \centering
    \includegraphics[width=0.49\textwidth]{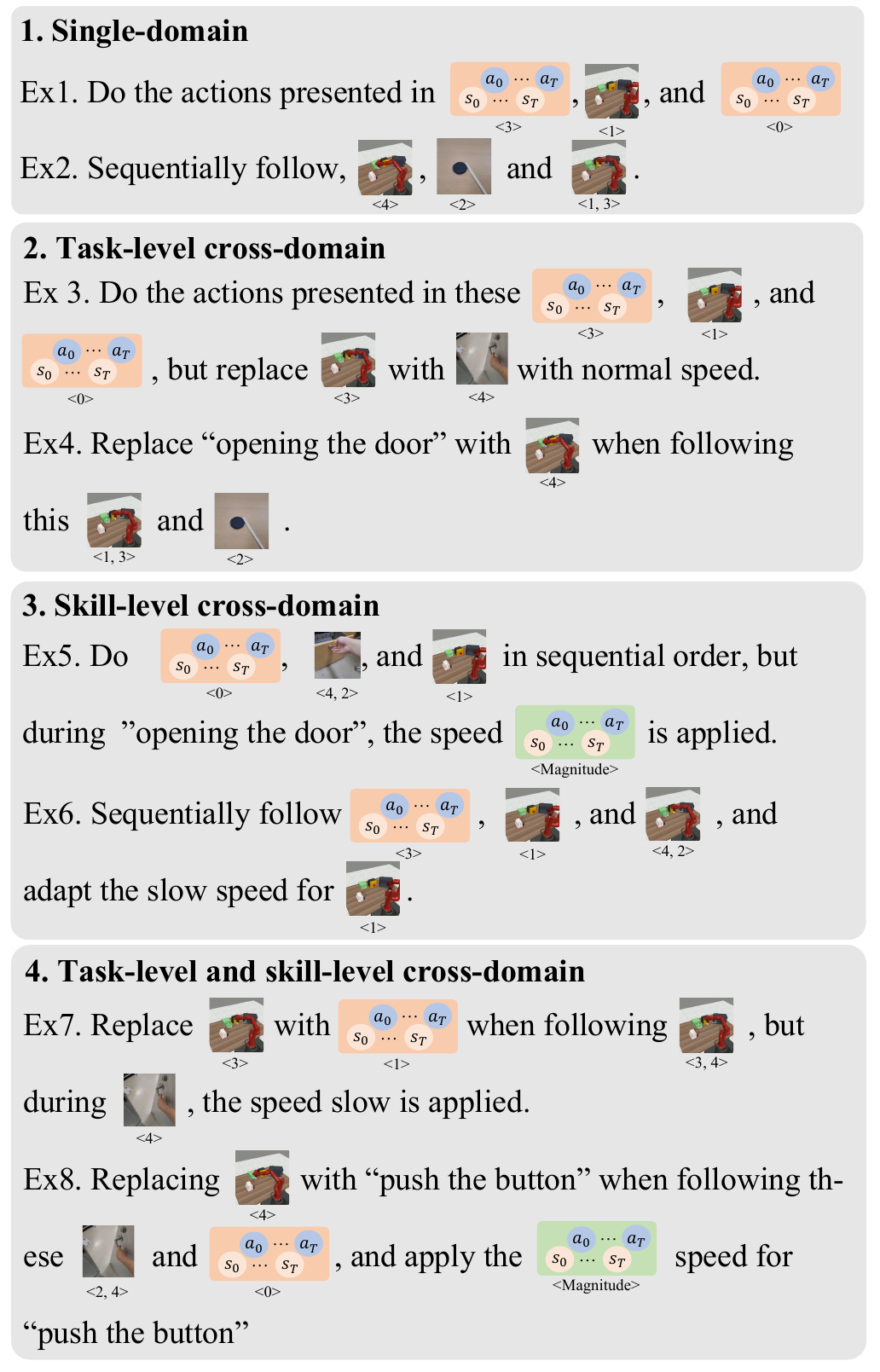}
    \caption{Multi-modal task prompts for Meta-World. The prompt comprises video, text, and sensor data. Certain snippets contain instructed cross-domain context, such as the speed requirement.}
    \label{appendix: fig: metaworld mm task prompts}
\end{figure}

Furthermore, we introduce time-varying parameters into the third axis of agents' actions to simulate a non-stationary environment as explained in Section~\ref{appendix: sec: complete results}. Examples of task prompts used in Franka Kitchen are in Listing~\ref{listing: kitchen prompts}.
Examples of task prompts used in Meta-World are in Listing~\ref{listing: metaworld prompts} and Figure~\ref{appendix: fig: metaworld mm task prompts}.

\begin{table}[h]
    \footnotesize
    \centering
    \begin{adjustbox}{width=0.48\textwidth}
    \begin{tabular}{c||c|c|c|c}
    \specialrule{.1em}{.05em}{.05em}
    \textbf{Domain} & \textbf{Slide puck} & \textbf{Close drawer} & \textbf{Push button} & \textbf{Open door}\\
    \specialrule{.1em}{.05em}{.05em}

    Slow & 166.08 & 162.23 & 177 & 374.35 \\ 
    Normal & 100.73 & 84.68 & 68.2 & 177 \\ 
    Fast & 75.8 & 56.62 & 53.1 & 104.34 \\ 
    \specialrule{.1em}{.05em}{.05em} 
    \end{tabular}
    \end{adjustbox}
    \caption{
    Performance in autonomous driving.
    }
    \label{appendix: tbl: metaworld speed}
\end{table}

\begin{lstlisting}[style=mypy, caption= Task prompt examples for Meta-World., label=listing: metaworld prompts]
# Single-domain
Ex1. "Do {SRC_TASK} in sequential order."   
Ex2. "Sequentially follow {SRC_TASK}."
# Task-level cross-domain
Ex3. "Do {SRC_TASK} in reverse order."  
Ex4. "Replace opening the door with closing the drawer when following {SRC_TASK}."
# Skill-level cross-domain
Ex5. "Do {SRC_TASK} in sequential order, but during opening the drawer, the speed slow is applied."  
Ex6. "Sequentially follow {SRC_TASK} and adapt the slow speed for pushing the puck into the goal."
# Task-level and Skill-level cross-domain
Ex7. "Replace opening the door with closing the drawer when following {SRC_TASK}, but during opening the drawer, the speed slow is applied."
Ex8. "Replacing opening the door with closing the drawer when following {SRC_TASK}, and adapt the slow speed for pushing the puck into the goal."
\end{lstlisting}

\subsection{RLBench}
\label{appendix: rlbench}
This environment simulates the cognitive robot manipulation scenario discussed in Table~\ref{exp: cognitive robot} of the main manuscript. In this environment, an 8-DoF robot arm needs to perform sequential subtasks, such as ``pick up red block'', and ``close the box''. 
To compute the affordance score of Saycan~\cite{saycan} in Table~\ref{exp: cognitive robot}, we collect semantically annotated datasets comprising 460 trajectories. We use a rule-based policy, which performs given skill-level instruction, for the evaluation of both SayCan and our framework.

In these scenarios, we explore three different command styles:
A \textbf{Summarized sentence} command that offers an abstract instruction rather than direct skill-level instruction for the agent; An \textbf{Abstract verbs} command that encompasses synonymous verbs to communicate actions; An \textbf{Embodiment} command that provides the current environment conditions and the constraints of the robots. 

For evaluations, we leverage advanced PLMs such as GPT3.5, GPT4, and PaLM in a zero-shot manner, in a similar way of~\cite{saycan}. Instead of fine-tuning, we engineer an augmented task prompt to contain beneficial information for semantic skill planning, which is presented to the PLMs with the respective commands. The examples of abstract commands and augmented engineered prompts are in Linsting~\ref{listing: cognitive robot commands}.
\begin{lstlisting}[style=mypy, caption= Engineered augmented prompt and abstract commands examples for RLBench., label=listing: cognitive robot commands]
# Prompt
Ex1. "You are a model trained to sequence tasks. You have a list of seven skills:

pick up red block
pick up blue block
pick up yellow block
close the box
open the box
place the block on the desk
place the block in the box

Your task is to sequence these skills in a logical order to complete a task. You can only use the skills in the list.
After all skills have been used, write `Finish'. Please list the skills in the order they should be performed, starting with `1. First skill 2. Second skill', and so on".

/* Summarized Sentence */
# Command examples
Ex2. "Place the red block into the closed box. Currently, Could you suggest an appropriate skill sequence to handle this situation?"
Ex3. "Move the blue block into the closed box. Currently, Could you suggest an appropriate skill sequence to handle this situation?"
Ex4. "Put the yellow block into the closed box. Currently, Could you suggest an appropriate skill sequence to handle this situation?"
/* Abstract verbs */
# Command examples
Ex5. "Given task is executed in the following skill sequence:
1. open the box
2. move the red block in the box
3. close the box
4. Finish
Currently, Could you suggest an appropriate skill sequence to handle this situation?"
Ex6. "Given task is executed in the following sequence:
1. move the red block on the top of the blue block to the desk
2. open the box
3. move the blue block in the box
4. close the box
5. Finish
Currently, Could you suggest an appropriate skill sequence to handle this situation?"
Ex7. "Given task is executed in the following sequence:
1. pick up {block_color_1} block
2. place the block on the desk
3. open the box
4. move the {block_color_2} block in the box
5. close the box
6. Finish
Currently, Could you suggest an appropriate skill sequence to handle this situation?"
/* Embodiment */
# Command examples
Ex8. "Given task is executed in the following skill sequence:
1. open the box
2. pick up red block
3. place the block in the box
4. close the box
5. Finish
Currently, a blue block is placed on top of the box lid and it cannot be opened. Could you suggest an appropriate skill sequence to handle this situation?"
Ex9. " Given task is executed in the following skill sequence:
1. open the box
2. pick up red block
3. place the block in the box
4. close the box
5. Finish
Currently, the box lid is already open. Could you suggest an appropriate skill sequence to handle this situation?"
Ex10. "Given task is executed in the following skill sequence:
1. open the box
2. pick up {block_color_1} block
3. place the block in the box
4. close the box
5. Finish
Currently, there is a {block_color_2} block is placed on top of the {block_color_1} block so you cannot pick it up. Could you suggest an appropriate skill sequence to handle this situation?"
\end{lstlisting}

\begin{figure*}[t]
    \centering
        \includegraphics[width=0.95\textwidth]{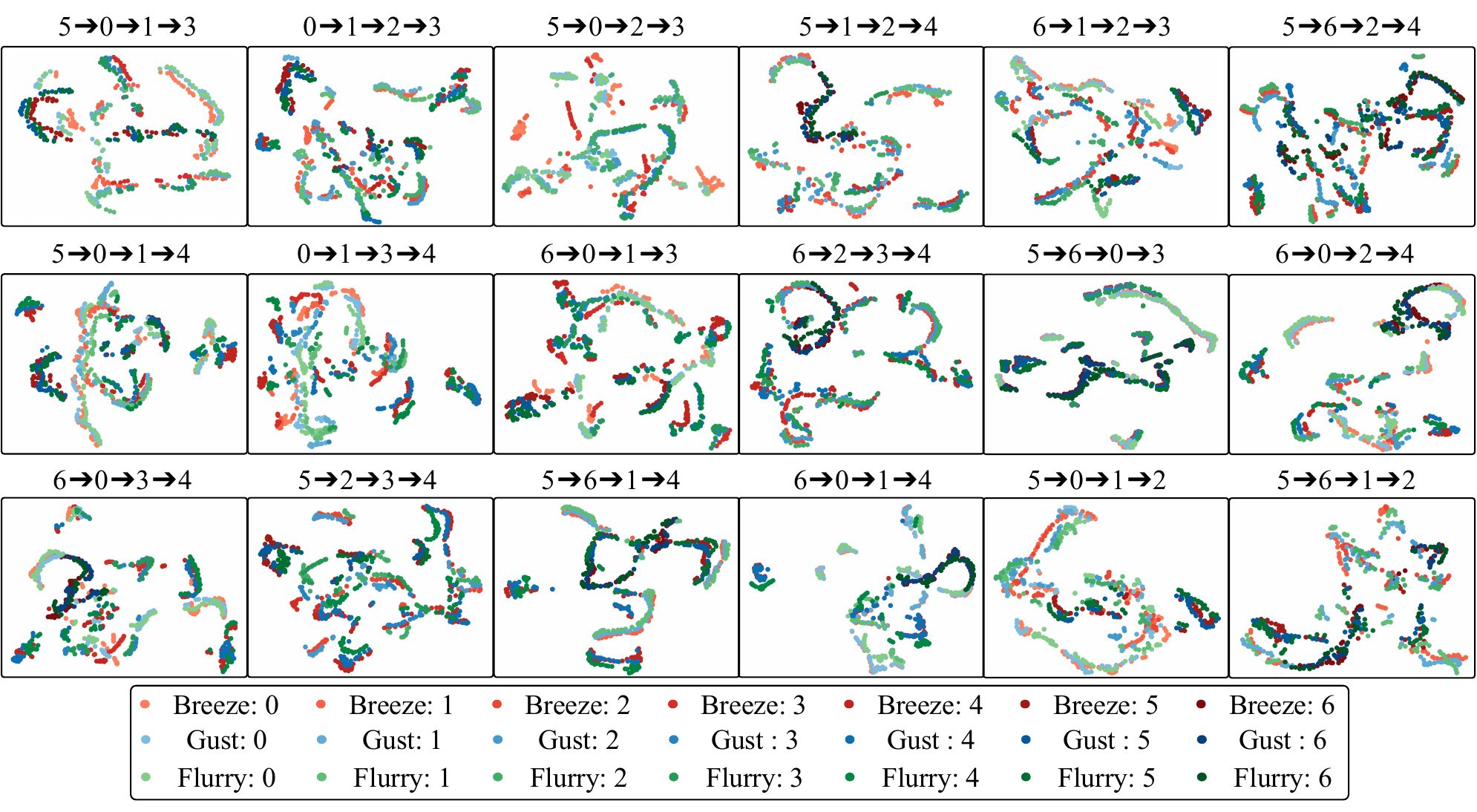}
    \caption{
    Cross-domain context affects the action distribution of adaptation policy: 
    In the figure, a numeric index corresponds to a specific semantic skill (for example, 0 corresponds to "open microwave"). 
    We use t-SNE to visualize the distribution of actions, which sequentially perform four semantic skills in a trajectory. 
    Each subfigure consists of three distinct trajectories, each of which performs the same order of semantic skill sequences but in different domain contexts. As observed, while the action sequences show a similar path when the semantics are the same, but perturbations occur based on the domain context.
    }
    \label{fig:methods}
\end{figure*}

\subsection{CARLA}
In this environment, the agent needs to adjust a 3-dimensional action for steering, throttle, and brake to reach a goal position via the shortest path. In this setting, we consider different car configurations (agent embodiments), such as sedan and truck, as distinct domain contexts. As described in Figure~\ref{fig:carla} of the main manuscript, the actual actions to be executed for skills, ``going straight'', ``turning left'', and ``turning right'', need to be adjusted according to the different car configurations.

Regarding the dataset generation, we collect 200 expert trajectories by implementing a waypoint-based expert policy for each car configuration.
The evaluation task involves the sequential execution of five skills: ``going straight'', ``turning right'', ``going straight'', ``turning left'', and ``going straight'', to reach the goal. 
The training dataset only contains trajectories composed of four or fewer skills; no trajectories involve all five skills in sequence. 
Consequently, the agent has to adapt to the target domain task by composing the skills learned during training, and aligning them with the domain context (car configuration).

\begin{lstlisting}[style=mypy, caption= Task prompt examples for CARLA., label=listing: carla prompts]
# Training task prompts
Ex1. "Go right, and then go straight to drive a truck."
Ex2. "Drive a truck go straight, and then go right."
Ex3. "Go straight, and then go right to drive a sedan."
Ex4. "Drive a sedan with go left, and then go straight."

# Evaluation task prompts
Ex5. "Go straight, then go right, then go straight, then go left, and then go straight, to drive a sedan."
Ex6. "Drive a truck with go straight, then go right, then go straight, then go left, and then go straight."

\end{lstlisting}

\section{Experiment Details}

\begin{algorithm}[t]
\caption{Evaluation of $\oursol$}
\begin{algorithmic}[1]
    \STATE Task prompt $\TP$
    \STATE Multi-modal skill enc. $\Phi_E$, skill sequence generator $\Phi_G$
    \STATE Skill boundary detector $\Phi_B$, context encoder $\Phi^{(g)}_C$
    \STATE Online context encoder $\Phi^{(o)}_C$, history buffer $\mathcal{B}$
    \STATE Behavior decoder $\pi$
    
    \nonumber \textit{/* Task interpretation */}
    \STATE $\eta \leftarrow \Phi_E(\TP)$ using~\eqref{eqn: multi-modal skill encoder}
    \STATE $\xi = (z_{j_1}, \cdots, z_{j_J}) \leftarrow \Phi_G(\eta)$ using~\eqref{sequence generator}
    \STATE $(\bar{z}_{j_1}, \cdots \bar{z}_{j_J}) \leftarrow \Phi^{(g)}_C(\eta, \xi)$ using~\eqref{eqn: context encoder}
    
    \nonumber \textit{/* Skill interpretation */}
    \STATE $t \leftarrow 0$, $i \leftarrow 0$, done $\leftarrow$ False
    \STATE $s_t \leftarrow$ env.reset(), $s_{t_0} \leftarrow s_0$, $\mathcal{B}$.append($s_t$)
    \WHILE{\textbf{not} done}
        \STATE $h_t \leftarrow \Phi^{(o)}_C(s_{t-H': t}, a_{t-H': t-1})$ using~\eqref{eqn: online context encoder}
        \STATE $a_t \leftarrow \pi(s_t, \bar{z}_{j_i}, h_t)$ using~\eqref{eqn: behavior decoder}
        \STATE $(s_{t+1}, \textnormal{done}) \leftarrow $ env.step($a_t$)
        \STATE $\mathcal{B}$.append($(s_{t+1}, a_t)$)
        \STATE  $m \leftarrow \Phi_B(s_t, z_{j_i}, s_{t_0})$ using ~\eqref{eqn: boundary detector},
        $t \leftarrow t + 1$
        \IF{$m = 1$}
            \STATE $s_{t_0} \leftarrow s_t$, $i \leftarrow i + 1$
        \ENDIF
    \ENDWHILE
\end{algorithmic}
\label{alg:eval}
\end{algorithm}

\subsection{Complete results}
\label{appendix: sec: complete results}
To implement a cross-domain scenario featuring temporal domain shift, we employ a non-stationary environment. In this setup, a time-varying parameter $w_t$ is added to specific axes of actions before an agent interacts with the environment; this interaction follows the transition probability
\begin{equation}
    s_{t+1} \sim \mathbb{P}(s_{t+1} | s_t, a_t + w_t),
\end{equation}
which the probability changes over time.

We explore four distinct non-stationary settings, each paired with three distinct instructed domain contexts (i.e., contexts that can be encapsulated in the task prompt or demonstration). This results in a total of twelve cross-domain contexts. These contexts are combined with nine unseen tasks (i.e., unseen order of subtasks) in Franka Kitchen and eight unseen tasks in Meta-World, forming 108 and 96 cross-domain environment scenarios, respectively.

Table~\ref{appendix exp: w/ temporal shift} presents the performance on a total of 12 cross-domain contexts, which are combinations of the three types of skill-level domain contexts and the four types of non-stationary contexts.
Due to their absence of history-aware model architecture, TP-BCZ and VCP are evaluated solely on instructed domain contexts, as described in Table~\ref{appendix exp: wo/ temporal shift}.

Below are the different types of time-varying parameters.
\label{appendix: exp_nonstationary_env}
\begin{itemize}
    \item
    Sin: 
    \begin{equation}
        w_t = 
        \begin{cases}
            0.6 \times \textnormal{sin} (\frac{8\pi \times t}{T}) \,\,\, \textnormal{if} \,\,\, u_t \leq 0.5 \\
            0.25 \times \textnormal{sin} (\frac{8\pi \times t^2}{T}) \,\,\, \textnormal{if} \,\,\, u_t > 0.5
        \end{cases}
    \end{equation}
    
    \item Bias(m)
    \begin{equation}
    w_t = 
    \begin{cases}
        -m + \textnormal{Unif}(0, 0.1) \,\,\, \textnormal{if} \,\,\, u_t \leq 0.25 \\
        m + \textnormal{Unif}(0, 0.1) \,\,\, \textnormal{if} \,\,\, u_t > 0.25 \\
    \end{cases}
    \end{equation}
    
    \item Bias(M)
    \begin{equation}
    w_t = 
    \begin{cases}
        -M + \textnormal{Unif}(0, 0.1) \,\,\, \textnormal{if} \,\,\, u_t \leq 0.25 \\
        M + \textnormal{Unif}(0, 0.1) \,\,\, \textnormal{if} \,\,\, u_t > 0.25 \\
    \end{cases}
    \end{equation}
    
    \item Zigzag
    \begin{equation}
    w_t = 
    \begin{cases}
        0.45 \,\,\, \textnormal{if} \,\,\, u_t \leq 0.5 \\
        -0.45 \,\,\, \textnormal{if} \,\,\, u_t > 0.5 \\
    \end{cases}
    \end{equation}
where $u_t \sim \textnormal{Unif}(0, 1)$ is a random value sampled at each timestep $t$.
\end{itemize}

\begin{table}[h]
    \footnotesize
    \centering
    \begin{adjustbox}{width=0.43\textwidth}
    \begin{tabular}{c|c||c|c|c}
    \specialrule{.1em}{.05em}{.05em}
    \textbf{Env.} & \textbf{Domain} & \textbf{VIMA} & \textbf{TP-GPT} & \textbf{SemTra} \\
    \specialrule{.1em}{.05em}{.05em}
    \multirow{12}{*}{\begin{tabular}{@{}c@{}}Franka \\ Kitchen\end{tabular}} 
    & breeze, sin & 39.82\% & 42.59\% & 59.26\% \\ 
    & breeze, bias(m) & 32.40\% & 42.59\% & 67.59\%\\
    & breeze, bias(M) & 35.18\% & 42.59\% & 64.81\% \\
    & breeze, zigzag & 35.18\% & 40.74\% & 50.00\% \\

    & gust, sin & 13.89\% & 15.74\% & 54.62\% \\
    & gust, bias(m) & 15.74\% & 14.82\% & 62.03\% \\ 
    & gust, bias(M) & 14.81\% & 10.19\% & 64.81\% \\ 
    & gust, zigzag & 10.19\% & 10.19\% & 58.33 \% \\

    & flurry, sin & 10.18\% & 0.00\% & 66.67\% \\
    & flurry, bias(m) & 7.41\% & 0.00\% & 66.67\% \\ 
    & flurry, bias(M) & 6.48\% & 0.00\% & 66.67\% \\ 
    & flurry, zigzag & 7.41\% & 1.86\% & 55.56\% \\
    \specialrule{.1em}{.05em}{.05em} 

    \multirow{12}{*}{\begin{tabular}{@{}c@{}}Meta- \\ World\end{tabular}} 
    & normal, sin & 4.17\% & 15.28\% & 73.61\% \\ 
    & normal, bias(m) & 4.17\% & 11.11\% & 81.94\% \\ 
    & normal, bias(M) & 1.39\% & 6.94\% & 80.56\% \\ 
    & normal, zigzag & 1.39\% & 12.50\% & 80.56\% \\ 

    & slow, sin & 1.39\% & 9.73\% & 56.94\%\\ 
    & slow, bias(m) & 0.00\% & 8.33\% & 58.33\% \\ 
    & slow, bias(M) & 0.00\% & 8.33\% & 62.50\% \\ 
    & slow, zigzag & 2.78\% & 11.11\% & 55.56\% \\ 

    & fast, sin & 0.00\% & 8.33\% & 86.11\% \\ 
    & fast, bias(m) & 0.00\% & 8.33\% & 86.11\% \\ 
    & fast, bias(M) & 0.00\% & 16.67\% & 79.17\% \\ 
    & fast, zigzag & 0.00\% & 11.11\% & 80.56\% \\ 
    \specialrule{.1em}{.05em}{.05em} 
    \end{tabular}
    \end{adjustbox}
    \caption{Performances in environments without temporal domain shifts.}
    \label{appendix exp: w/ temporal shift}
\end{table} 

\begin{table}[h]
    \footnotesize
    \centering
    \begin{adjustbox}{width=0.48\textwidth}
    \begin{tabular}{c|c||c|c|c|c|c}
    \specialrule{.1em}{.05em}{.05em}
    \textbf{Env.} & \textbf{Domain} & \textbf{VIMA} & \textbf{TP-BCZ} & \textbf{TP-GPT} & \textbf{VCP} & \textbf{SemTra} \\
    \specialrule{.1em}{.05em}{.05em}

    \multirow{9}{*}{\begin{tabular}{@{}c@{}}Franka \\ Kitchen\end{tabular}} 
    & seq, breeze & 48.15\% & 22.22\% & 59.26\% & 43.52\% & 73.15\% \\ 
    & rev, breeze & 46.30\% & 14.82\% & 50.93\% & 62.97\% & 71.30\% \\ 
    & rep, breeze & 30.95\% & 15.48\% & 48.81\% & 26.19\% & 64.29\% \\ 

    & seq, gust & 38.89\% & 23.15\% & 17.59\% & 15.74\% & 69.45\% \\ 
    & rev, gust & 38.89\% & 22.22\% & 26.86\% & 20.37\% & 71.30\% \\ 
    & rep, gust & 23.81\% & 11.90\% & 25.00\% & 27.39\% & 64.29\% \\  

    & seq, flurry & 8.33\% & 19.45\% & 1.86\% & 16.67\% & 60.19\% \\ 
    & rev, flurry & 10.19\% & 25.30\% & 2.78\% & 18.52\% & 62.04\% \\ 
    & rep, flurry & 3.57\% & 16.67\% & 2.39\% & 38.10\% & 58.33\% \\ 
    \specialrule{.1em}{.05em}{.05em} 

    \multirow{9}{*}{\begin{tabular}{@{}c@{}}Meta- \\ World\end{tabular}} 
    & seq, stand. & 16.67\% & 15.28\% & 22.22\% & 20.93\% & 80.56\% \\ 
    & rev, stand. & 5.56\% & 8.33\% & 22.22\% & 8.33\% & 76.39\% \\ 
    & rep, stand. & 8.89\% & 8.89\% & 20.00\% & 2.22\% & 80.00\% \\ 

    & seq, slow & 5.56\% & 12.50\% & 8.33\% & 4.17\% & 94.44\% \\ 
    & rev, slow. & 2.78\% & 12.50\% & 5.56\% & 9.72\% & 79.17\% \\ 
    & rep, slow. & 4.45\% & 11.11\% & 13.33\% & 0.00\% & 97.78\% \\  

    & seq, fast. & 12.50\% & 0.00\% & 8.33\% & 11.11\% & 73.61\% \\ 
    & rev, fast. & 5.56\% & 0.00\% & 6.95\% & 15.28\% & 72.22\% \\ 
    & rep, fast. & 6.67\% & 2.22\% & 11.11\% & 4.44\% & 60.00\% \\  
    \specialrule{.1em}{.05em}{.05em} 
    
    \end{tabular}
    \end{adjustbox}
    \caption{Performances in environments with temporal domain shifts.}
    \label{appendix exp: wo/ temporal shift}
\end{table} 

\subsection{Visual cross-domain evaluation}
\begin{figure}[h]
    \centering
    \includegraphics[width=0.48\textwidth]{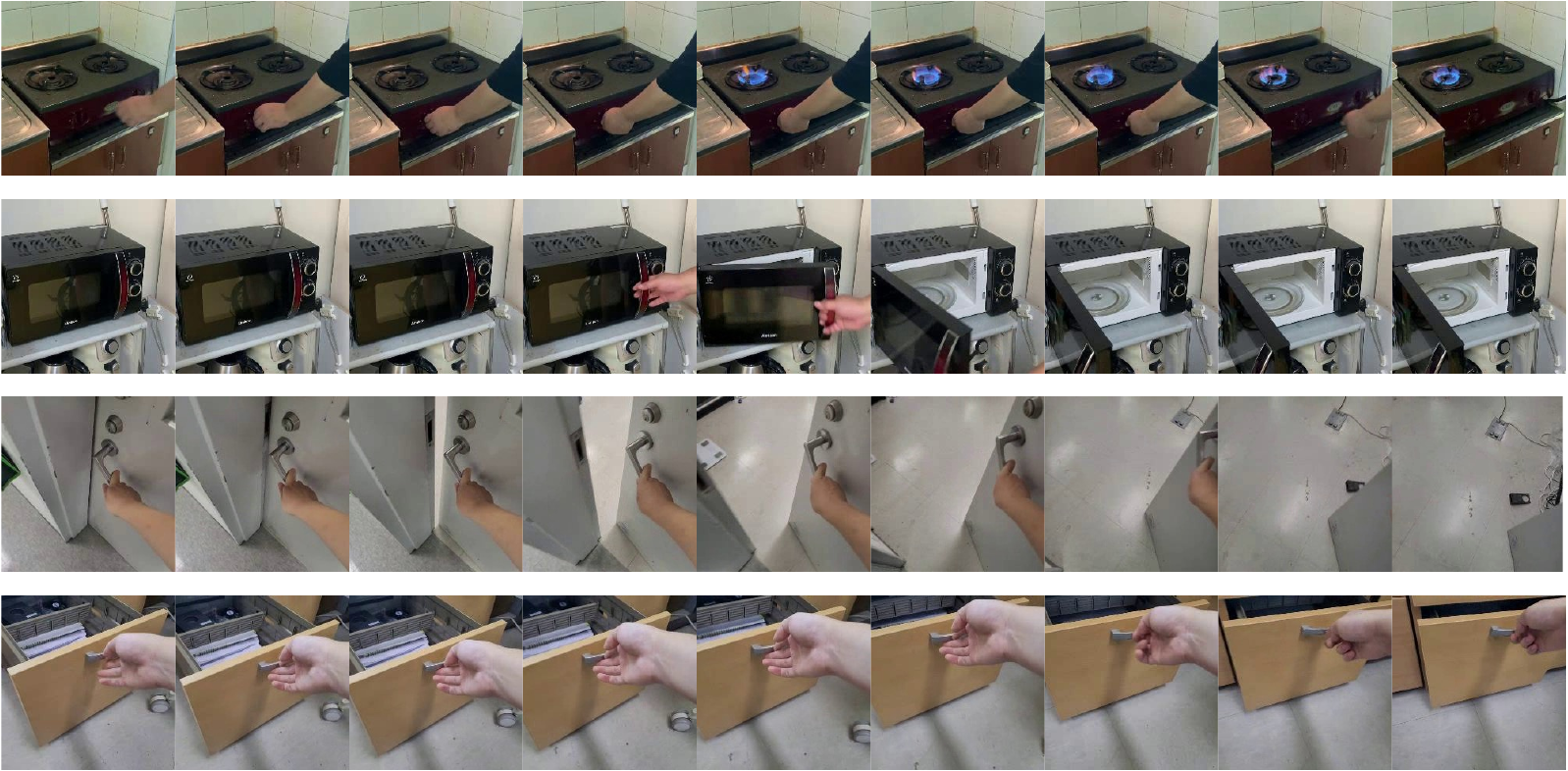}
    \caption{
    Examples of real-world videos used for visual cross-domain scenarios. From top to bottom, the video represents ``turn on burner'', ``open microwave'', ``open door'', and ``close drawer''. Each video snippet constitutes a multi-modal task prompt.
    }
    \label{appendix: fig: real world videos}
\end{figure}

We record real-world video demonstrations of subtasks that construct Franka Kitchen and Meta-World environments at a resolution of $700 \times 700$ in residential kitchens and offices, respectively. Then, to generate a noisy human demonstration, we add a Gaussian noise which is sampled from the distribution of $\mathcal{N}(0, 1)$. Finally, to generate an illustrative (animation) demonstration we use AnimeGAN~\cite{chen2020animegan}, which transfers the video style from real world to animation. 
These videos are videos were then cropped to a size of 224x224 and used as input to V-CLIP~\cite{v-clip}. Examples of the real-world videos are showcased in Figure~\ref{appendix: fig: real world videos}.

\subsubsection{Video encoder tuning for human demonstrations.}
To analyze the effect of fine-tuning the VLM-based video skill encoder in~\eqref{eqn: video skill encoder}, we evaluate our framework with a task prompt involving noisy human-level demonstrations in various scenarios: demonstrations without noise (wo/ noise in the Demo column of Table~\ref{exp: human_demo}), noisy demos, and illustrative (animation) videos. 
Noise-free demos are accurately translated to semantic skills due to the domain-invariant semantic knowledge of the pre-trained VLM (wo/ F.T in the column name). Fine-tuning the skill encoder (w/ F.T) enhances the skill prediction when a demonstration is semantically annotated. Since the skill sequence generator is trained in the language semantic space, it generates suitable skill sequences, resulting in only a minor performance drop.
\begin{table}[h]
    \footnotesize
    \centering
    \begin{adjustbox}{width=0.48\textwidth}
           \begin{tabular}{c||c|cc|cc}
                \specialrule{.1em}{.05em}{.05em}
                \multirow{2}{*}{\textbf{Env}} & \multirow{2}{*}{\textbf{Domain}} & 
                \multicolumn{2}{c}{\textbf{Skill prediction}} & 
                \multicolumn{2}{c}{\textbf{Adaptation}} \\
                && wo/ F.T & w/ F.T & wo/ F.T & w/ F.T \\
                \specialrule{.1em}{.05em}{.05em} 
                \multirow{3}{*}{\begin{tabular}{@{}c@{}}Franka \\ Kitchen\end{tabular}} 
                & wo/ noise & 94.44\% & 97.22\% & 57.41\% & 70.37\% \\ 
                \cline{2-6}
                & w/ noise & 86.11\% & 94.44\% & 67.59\% & 69.44\% \\ 
                & Illust. & 75.00\% & 94.44\% & 37.04\% & 62.97\% \\ 
                \specialrule{.1em}{.05em}{.05em}
                \multirow{3}{*}{\begin{tabular}{@{}c@{}}Meta- \\ World\end{tabular}} 
                & wo/ noise & 100.0\% & 100.0\% & 75.00\% & 75.00\% \\  
                \cline{2-6}
                & w/ noise & 83.33\% & 95.83\% & 55.56\% & 69.44\% \\  
                & Illust. & 79.17\% & 91.67\% & 40.28\% & 69.50\% \\ 
                \specialrule{.1em}{.05em}{.05em} 
                \end{tabular}
                \end{adjustbox}
    \caption{Fine-tuning for human demonstrations.}
    \label{exp: human_demo}
\end{table}
\end{document}